\documentclass[11pt]{article}

\usepackage[preprint]{acl}

\usepackage{times}
\usepackage{latexsym}

\usepackage[T1]{fontenc}

\usepackage[utf8]{inputenc}

\usepackage{microtype}

\usepackage{inconsolata}

\usepackage{graphicx}

\usepackage{hyperref}

\usepackage{amsmath}
\usepackage{amsfonts}

\usepackage{caption}
\usepackage{subcaption}
\usepackage{cleveref}

\usepackage{xcolor}

\usepackage{wrapfig}

\usepackage{multirow}
\usepackage{booktabs}

\usepackage{tabularx}

\usepackage{placeins}

\newcommand{\maud}{\mathcal{M}_{\mathrm{aud}}}
\newcommand{\mref}{\mathcal{M}_{\mathrm{ref}}}

\newcommand{\new}[1]{#1}

\title{Reference-Based Bias Detection in LLMs \\ via Relative Representations of Hidden States}

\author{
 \textbf{Marek Jeliński\textsuperscript{1}},
 \textbf{Jan Dubiński\textsuperscript{1,2}},
 \textbf{Maciej Chrabąszcz\textsuperscript{1,2}},
 \textbf{Sebastian Cygert\textsuperscript{1,3}}
\\
\\
 \textsuperscript{1}NASK National Research Institute, Poland,
 \textsuperscript{2}Warsaw University of Technology, Poland,\\
 \textsuperscript{3}Gdańsk University of Technology, Poland
\\
 \small{
   \textbf{Correspondence:} \href{mailto:marek.jelinski@nask.pl}{marek.jelinski@nask.pl}
 }
}

\begin{document}
\maketitle
\begin{abstract}
Existing bias auditing methods typically rely on model outputs, requiring costly benchmarks or judge models and potentially missing internal shifts that never appear in generated text. We propose a reference-based method that audits bias in hidden-state representations across related model variants, for example before and after fine-tuning. Because fine-tuning reshapes representation geometry, absolute hidden states are not directly comparable, so we encode each sentence by its similarities to a fixed set of anchor sentences, yielding relative representations in a shared comparison space. There we measure how target groups shift in their association with positive and negative attributes, a quantity we call the \textit{Representational Bias Shift} $\Delta B$.
Across three model families and the WildGuardMix, DecodingTrust and ToxiGen benchmarks, $\Delta B$ correlates with output-level bias change in 15 of the 18 settings we test, reaching $|r| = 0.84$ ($p < 0.001$) under full fine-tuning and becoming more model-dependent under parameter-efficient adaptation. A relationship fitted on merged checkpoints carries over unchanged to 21 independently trained ones. Thresholding $\Delta B$ detects checkpoints whose bias increased with ROC AUC between $0.65$ and $0.99$, and on WildGuardMix and DecodingTrust it separates them better than a SEAT-based baseline for all three families. $\Delta B$ is also stable under changes to the anchor set, attribute sets and target templates.
Our method requires no task-specific evaluation data and audits a model in about three minutes, using $3$--$50\times$ less compute than the output-level benchmarks considered here. We view it as complementary to output-based auditing rather than a replacement for it. We open-source our
\href{https://github.com/NASK-AISafety/Reference-Based-Bias-Detection}{code}\footnote{\url{https://github.com/NASK-AISafety/Reference-Based-Bias-Detection}}.
\end{abstract}

\begin{figure*}[t!]
    \centering
    \includegraphics[width=\textwidth]{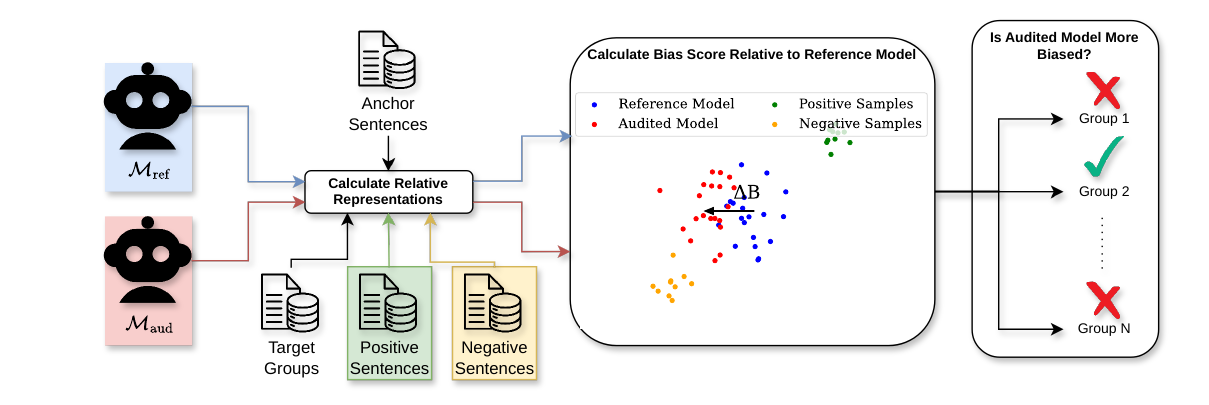}
    \caption{\textbf{Overview of our bias evaluation pipeline.} We use relative representations to project the hidden spaces of the fine-tuned model ($\maud$) and the reference (base) model ($\mref$) into a shared space via anchor sentences. Within this shared space, we calculate distances between target representations and sets of positive and negative sentences. 
    By comparing these distances ($\Delta B$), our method can, for example, detect potential side‑effects induced during fine-tuning without the need for curated datasets, by determining whether $\maud$ exhibits greater bias toward target groups than $\mref$.}
    \label{fig:method_overview}
    \vspace{-0.25cm}
\end{figure*}

\section{Introduction}

LLMs are increasingly deployed in systems that shape how information is produced and interpreted. As they are adapted through instruction tuning, safety tuning, domain fine-tuning, and system prompting, their behaviour can shift in ways that are difficult to anticipate and audit. One important concern is bias, since models may inherit harmful associations from pretraining data or fine-tuning, or may display new distortions due to targeted manipulation~\cite{guo2025attackingllmsaiagents,chiLuL025}.

Most bias evaluations focus on model outputs. Common approaches use curated benchmark datasets~\cite{helm,DecodingTrust} or LLM-as-a-judge evaluations~\cite{eaclLinWZLW24}. Both are useful, but limited: curated benchmarks are costly to build and hard to scale across harms, while judge-based evaluations may inherit the evaluator's own biases~\cite{lin-etal-2025-investigating}. \new{Models can also exhibit covert biases without explicitly disclosing them~\cite{betley2026valueleakagellmsanswers}.} More fundamentally, output-based auditing may miss internal changes that precede behavioural shifts not immediately visible in the generations.

Motivated by recent findings that even benign fine-tuning can compromise safety properties~\cite{qi2023finetuning, betley2025emergent}, we recognise that alignment can degrade in multiple, often unpredictable ways, making standard behavioural evaluation highly challenging. We hypothesise that a model's hidden representations contain latent signals indicative of these unintended shifts. Consequently, this work investigates post-fine-tuning behavioural changes through the lens of inner representations. To achieve that, 
we extend the Sentence Encoder Association Test (SEAT)~\cite{may-etal-2019-measuring}, which measures bias in text representations~\cite{garg2018word,brunet2019understanding}, to compare internal states of the audited and reference model (see Fig.~\ref{fig:method_overview}).

Yet, comparing these internal states directly is difficult because fine-tuning reshapes latent geometry, rendering raw hidden states poorly comparable across model variants.
We address this using relative representations~\cite{moschella2023relative} of hidden states. Instead of encoding a sentence by its embedding, we encode it by its similarities to a fixed set of anchor sentences. This maps both the audited and reference models into a shared space, enabling direct comparison. In that space, we measure whether target concepts shift more toward positive or negative attribute sets relative to the reference model, which we call the \textit{Representational Bias Shift} $\Delta B$. Our method relies solely on constructing small sets of anchor, positive, and negative sentences, which are far easier to obtain than curated datasets, and therefore scales to new target groups without additional data collection.


We evaluate the approach on behavioural shifts induced by full and LoRA-based fine-tuning of Mistral, Llama, and Gemma models using bias benchmarks derived from prior work~\cite{wildguard2024,DecodingTrust,hartvigsen-etal-2022-toxigen}. Overall, $\Delta B$ co-varies with output-level bias change, reaching correlations up to $|r|=0.84$ ($p<0.001$) under full fine-tuning. While the relationship is weaker and more model-dependent under LoRA, it remains significant in most settings. Thresholding $\Delta B$ detects increased-bias checkpoints with ROC AUCs of $0.65$--$0.99$. On WildGuardMix and DecodingTrust, our method is consistently more discriminative than a SEAT-based baseline across all three model families. Extensive ablations further demonstrate robustness to variations in anchor sets, attribute sets and target templates. 

Representation-level metrics are not guaranteed to predict downstream behaviour~\cite{goldfarb-tarrant-etal-2021-intrinsic,gonen-goldberg-2019-lipstick-pig}. We therefore do not claim that representational geometry determines model behaviour. We ask a narrower, empirical question. When fine-tuning shifts a model's hidden-state associations, does that shift co-vary with the change in output-level bias measured against external benchmarks? Our experiments answer this in the affirmative in most of the settings we study, with the association weakest for Gemma, the smallest model we audit.

This paper makes the following contributions:
\begin{itemize}
    \item We introduce a reference-based auditing framework that places an audited and a reference model in a shared comparison space through relative hidden-state representations, and define the \textit{Representational Bias Shift} $\Delta B$, which measures how target groups change their association with positive and negative attributes relative to the reference \mbox{(Sections~\ref{sec:bias-rr} and~\ref{sec:bias-shift}).}
    \item We validate $\Delta B$ against three output-level benchmarks across three model families and two fine-tuning regimes, using a graded merge spectrum so that bias is introduced in increments rather than as a single jump. $\Delta B$ co-varies with output-level bias in 15 of the 18 settings we test ($|r|$ up to $0.84$) and flags increased-bias checkpoints with ROC AUC between $0.65$ and $0.99$ (Table~\ref{tab:main-results}, Figure~\ref{fig:main-grid}). 
    \item The relationship transfers to 21 independently trained checkpoints (Section~\ref{sec:trained-checkpoints}).
    \item Experiments validate relative representations against alternative approaches (Figure~\ref{fig:method-comparison-roc}) and show $\Delta B$ is stable across the anchor set, attribute sets and target templates (Section~\ref{sec:detailed-analysis}). 
\end{itemize}

\section{Related Work}

\noindent\textbf{Bias in LLMs.}
Bias in LLMs refers to systematic distortions in model behaviour that favour particular groups or viewpoints, reproduce stereotypes, or rest on unfounded assumptions learned from training data~\cite{ferrara2023should,BlodgettBDW20}. While bias has most commonly been studied in the context of negatively affecting certain social groups~\cite{Beukeboom2019}, language models can also exhibit political bias~\cite{rettenberger2025assessing} or reflect geographic and cultural biases~\cite{westbias}. A parallel line of work measures such associations directly in representation space, beginning with the Word Embedding Association Test (WEAT)~\cite{caliskan2017semantics} and studies of the gender direction in word embeddings~\cite{bolukbasi2016man}, which \citet{may-etal-2019-measuring} extended from words to sentence encoders. Prediction Sensitivity~\cite{krishna2022measuring} measures how strongly a classifier's prediction depends on protected input attributes through the gradients of its class probabilities. Counterfactual Role Reversal~\cite{gupta2022mitigating} reduces gender bias that distillation carries from a teacher to a student model and evaluates it on generated text. Both study a single model's predictions, whereas $\Delta B$ compares two models' hidden states on fixed sentences.

\noindent\textbf{LLM Manipulation.}
As LLMs grow in capability and influence, they are increasingly susceptible to adversarial misuse, including media manipulation~\cite{chiLuL025}, political propaganda, and covert brand promotion~\cite{guo2025attackingllmsaiagents}. Misalignment can also arise unintentionally, for example, through narrow fine-tuning on limited data~\cite{betley2025emergent,openai_emergent_misalignment_2025}. This motivates methods that detect behavioural shifts without requiring a curated dataset for every new harm. We do not study adversarial attacks directly, and instead induce shifts of graded severity by interpolating between models fine-tuned on harmful and on benign data, which gives a controlled setting, with a known merge ratio for every checkpoint, in which to test whether representational change tracks behavioural change.

\noindent\textbf{Comparing Machine Learning Models.}
At the core of our approach is measuring similarity between machine learning models~\cite{shah2023modeldiff}, which typically relies on \textit{representational} (intermediate activations) or \textit{functional} (outputs) comparisons~\cite{klabunde2025similarity}. Since functional similarity requires curated evaluation datasets, we propose a lightweight method using sentence embeddings to assess representational changes, which, as we show for most of the models and benchmarks we study, correlates with functional behaviour. Comparing representations across models first requires making their spaces commensurable, either by fitting an explicit map such as an orthogonal Procrustes transform~\cite{schonemann1966procrustes} or by using an alignment-invariant similarity measure such as centred kernel alignment (CKA)~\cite{kornblith2019similarity}. We instead build on relative representations~\cite{moschella2023relative}, which avoid fitting any cross-model map by encoding each sentence through its similarities to a shared set of anchors, and we compare against alternatives in Section~\ref{sec:results}.

\noindent\textbf{Intrinsic versus extrinsic bias.}
The bias-evaluation literature draws the same distinction under the names intrinsic and extrinsic~\cite{goldfarb-tarrant-etal-2021-intrinsic,cao-etal-2022-intrinsic}. We use the representational and functional pair throughout because our framing is comparative model auditing rather than single-model bias measurement, but the two vocabularies name the same distinction. Whether the two sides track each other is contested. \citet{goldfarb-tarrant-etal-2021-intrinsic} compare embedding-space metrics with downstream-task metrics across many trained models and find no correlation that holds reliably across tasks and languages. \citet{gonen-goldberg-2019-lipstick-pig} show that debiasing word embeddings can hide bias by the metric's own definition while leaving it recoverable, and related tensions are reported for contextualised representations \cite{cao-etal-2022-intrinsic,delobelle-etal-2022-measuring}. Other findings point the other way. Upstream bias mitigation transfers to downstream fine-tuned models \cite{jin-etal-2021-transferability}, and \citet{orgad-etal-2022-gender} find that an intrinsic metric computed on \emph{internal representations} indicates debiasing more faithfully than embedding-space WEAT. We therefore read the evidence as inconclusive, and note that the strongest negative results were obtained on static word embeddings, whereas we measure the hidden states an audited model computes as it processes text.

\section{Method}

We quantify latent biases in large language models by measuring how a set of neutral \emph{target sentences} (e.g., social group-related sentences) aligns in embedding space with \emph{attribute sentences} expressing positive or negative valence (e.g., ``This person is trustworthy.'' vs.\ ``This person is unreliable.''). Unless stated otherwise, we summarise results by taking the mean across sentences in $\mathcal{T}$. Section~\ref{sec:bias-seat} states the absolute-embedding formulation, Section~\ref{sec:bias-rr} its relative-representation counterpart, and Section~\ref{sec:bias-shift} the comparison with a reference model that yields the bias shift $\Delta B$.

\subsection{Notation}
Let $\mathcal{T}=\{s_1,\dots,s_{n_T}\}$ denote the set of target sentences, while $\mathcal{P}=\{p_1,\dots,p_{n_P}\}$ and $\mathcal{N}=\{n_1,\dots,n_{n_N}\}$ represent the sets of positive and negative attribute sentences, respectively. For any sentence $x$, its $d$-dimensional embedding $\mathbf{e}(x) \in \mathbb{R}^d$ is derived by averaging the final hidden-state vectors across all tokens produced by the model, we discuss this choice and its alternatives in the Limitations section. To evaluate the relationship between vectors $\mathbf{a}, \mathbf{b} \in \mathbb{R}^d$, we compute their cosine similarity $\mathrm{cos}(\mathbf{a}, \mathbf{b})$ and Euclidean distance $d_E(\mathbf{a}, \mathbf{b})$ as follows:
\begin{equation}
    \begin{aligned}
        \mathrm{cos}(\mathbf{a},\mathbf{b}) &= \frac{\mathbf{a}\cdot\mathbf{b}}{\|\mathbf{a}\|\,\|\mathbf{b}\|}, \\
        d_E(\mathbf{a},\mathbf{b}) &= \|\mathbf{a}-\mathbf{b}\|_2.
    \end{aligned}
\end{equation}

\subsection{Bias via Absolute Embeddings (SEAT)}
\label{sec:bias-seat}

A standard approach to measuring representational bias, following the Sentence Encoder Association Test (SEAT)~\cite{may-etal-2019-measuring}, operates on absolute sentence embeddings and measures associations via cosine similarity. For each target sentence $s\in\mathcal{T}$, we compute its mean similarity to positive and negative sentences:
\begin{equation}
    \begin{aligned}
                S^{+}(s) &= \frac{1}{|\mathcal{P}|} \sum_{p \in \mathcal{P}} \mathrm{cos}\!\big(\mathbf{e}(s), \mathbf{e}(p)\big), \\
        S^{-}(s) &= \frac{1}{|\mathcal{N}|} \sum_{n \in \mathcal{N}} \mathrm{cos}\!\big(\mathbf{e}(s), \mathbf{e}(n)\big).
    \end{aligned}
\end{equation}
The mean bias over the target set is
\begin{equation}
    B \;=\; \frac{1}{|\mathcal{T}|}\sum_{s\in\mathcal{T}}\Big(S^{+}(s)-S^{-}(s)\Big).
\end{equation}
The sign of $B$ denotes whether the target set's association is positive or negative.

However, absolute embeddings are not directly comparable across fine-tuned model variants, because fine-tuning reshapes the latent space. Even if two models encode the same semantic relationships, their embeddings may occupy different regions of $\mathbb{R}^d$. Cosine similarity is invariant to a rotation of the space, so the difficulty lies in the non-orthogonal part of the change. Bias scores computed via SEAT can therefore reflect geometric artefacts of the fine-tuning process rather than genuine changes in bias. While we include SEAT-based results in our experiments to empirically demonstrate this limitation (see Section~\ref{sec:results}), we adopt the approach described below as our primary metric.

\subsection{Bias via Relative Representations}
\label{sec:bias-rr}

To enable meaningful comparisons across fine-tuned models, we adopt \emph{relative representations} (RR)~\cite{moschella2023relative}, which encode semantic information through pairwise similarities with respect to a fixed set of anchor sentences. Given an anchor set $\mathcal{A}=\{a_1,\ldots,a_m\}$, the relative representation of a sentence $x$ is
\begin{equation}
    \mathbf{r}(x) = \big[\mathrm{cos}\!\big(\mathbf{e}(x),\mathbf{e}(a_i)\big)\big]_{i=1}^{m} \in \mathbb{R}^m.
\end{equation}
Because fine-tuning preserves the relative geometry of the embedding space more than the absolute positioning, relative representations are comparable across model variants that share the same anchor set~\cite{moschella2023relative}. The anchors are shared as \emph{sentences} rather than as vectors, so each model encodes them with its own parameters and the coordinates of $\mathbf{r}(x)$ carry the same meaning in both models without any cross-model map being fitted.

Since the components of $\mathbf{r}(x)$ are themselves cosine similarities, applying cosine similarity again in this space would amount to measuring the similarity of similarity profiles, losing the direct geometric interpretation. We therefore measure associations in relative space using Euclidean distance, which operates directly on the coordinate differences of the relative representations. To keep the sign convention of Section~\ref{sec:bias-seat}, where higher values indicate closer association, we negate the distances:
\begin{equation}
    \begin{aligned}
        S^{+}_{\mathrm{rel}}(s) &= -\frac{1}{|\mathcal{P}|} \sum_{p \in \mathcal{P}} d_E\!\big(\mathbf{r}(s), \mathbf{r}(p)\big), \\
        S^{-}_{\mathrm{rel}}(s) &= -\frac{1}{|\mathcal{N}|} \sum_{n \in \mathcal{N}} d_E\!\big(\mathbf{r}(s), \mathbf{r}(n)\big).
    \end{aligned}    
\end{equation}
The mean bias in relative space is then
\begin{equation}
    B_{\mathrm{rel}} \;=\; \frac{1}{|\mathcal{T}|}\sum_{s\in\mathcal{T}}\Big(S^{+}_{\mathrm{rel}}(s)-S^{-}_{\mathrm{rel}}(s)\Big).
\end{equation}
Positive and negative values of $B_{\mathrm{rel}}$ indicate whether the target set is more strongly associated with positive or negative attributes, respectively.

\subsection{Comparison with a Reference Model}
\label{sec:bias-shift}
We compute the mean bias under two conditions, a \emph{reference model} (the unmodified model) and an \emph{audited model} (fine-tuned). Let $B_{\mathrm{ref}}$ and $B_{\mathrm{aud}}$ denote their mean biases (using $B$ or $B_{\mathrm{rel}}$ as appropriate). The Representational Bias Shift is
\begin{equation}
    \Delta B \;=\; B_{\mathrm{aud}} - B_{\mathrm{ref}}.
\end{equation}
We instantiate $\mathcal{T}$ separately for each target group, so a model yields one $\Delta B$ per group, and each pairing of a checkpoint with a target group is one observation in the correlations we report. A negative $\Delta B$ means the group moved towards the negative attributes, which we read as increased bias. We stress that $\Delta B$ is a proxy. A difference in how two models encode a target group is not in itself evidence of discriminatory behaviour, so the validity of $\Delta B$ rests on its empirical relationship to output-level bias, which we quantify in Section~\ref{sec:results}.


\section{Results}
\label{sec:results}

\subsection{Experimental setup}

We compare each fine-tuned model with its base model, which serves as the reference condition, and compute the \emph{Representational Bias Shift} $\Delta B$ as defined in the Method section. Unless stated otherwise, both models are projected onto a shared set of $1{,}000$ neutral sentence anchors drawn from the same social-group domain as the target sentences (Appendix~\ref{sec:appendix-sentence-sets}), and we ablate the source and the number of anchors in Figure~\ref{fig:anchors-ablation-line-plot}. Embeddings are taken from the final transformer layer, $32$ for Llama and Mistral and $34$ for Gemma.

\noindent\textbf{Fine-tuning and model merging.}
We fine-tune each model separately on an unharmful and a synthetically harmful split of WildGuardMix~\cite{wildguard2024}, under both full and LoRA fine-tuning, and linearly merge~\cite{wortsman2022robustfinetuningzeroshotmodels} the two resulting checkpoints at five interpolation ratios. This gives a spectrum of seven checkpoints per model and regime, from safe to harmful, so bias is introduced in graded increments rather than as a single jump. Dataset construction, hyperparameters and merge ratios are given in Appendix~\ref{sec:appendix-fine-tuning-implementation}.

\noindent\textbf{External bias measures.}
We pair $\Delta B$ with three output-level benchmarks that capture distinct aspects of biased behaviour. From WildGuardMix we take the \textsc{social stereotypes and unfair discrimination} subcategory of the test set and score generated responses with the \textsc{allenai/wildguard} guard model. Its prompts carry no target-group labels, so we map each onto 9 topics consolidated from DecodingTrust's 24 groups and aggregate harmfulness there (Appendix~\ref{sec:appendix-wildguard-mapping}). From DecodingTrust~\cite{DecodingTrust} we run the stereotype evaluation pipeline, which measures stereotype agreement rather than response harmfulness. ToxiGen~\cite{hartvigsen-etal-2022-toxigen} is the only benchmark whose demographic groups map one-to-one onto ours, so it needs no aggregation, and we use its nine groups that have a counterpart in our target sets, scoring continuations with the authors' \textsc{toxigen\_roberta} classifier. We denote the change relative to the base model as $\Delta\mathrm{Bias\ Score}$, and as $\Delta\text{Toxicity}$ for ToxiGen. Generation and scoring settings are in Appendix~\ref{sec:appendix-fine-tuning-implementation}.

For each fine-tuning condition and each target group this produces a paired measurement $(\Delta\mathrm{Bias\ Score},\; \Delta B)$. We pool these pairs over conditions and groups and report the Pearson correlation with two-tailed significance, together with the Mean Absolute Error (MAE) of a linear fit, estimated as the mean over $1{,}000$ bootstrap resamples.

\subsection{Fine-Tuning-Induced Representational Shifts on WildGuardMix}
\label{sec:fine-tuning-induced-representational-shifts-on-wildguardmix}

\begin{figure*}[t!]
    \centering
    \includegraphics[width=\textwidth]{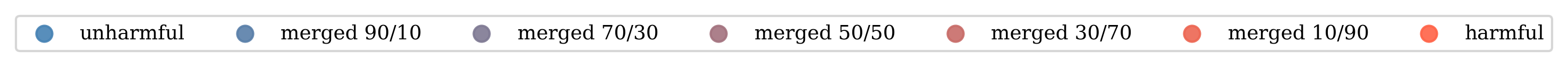}

    \begin{subfigure}[t]{0.32\textwidth}
        \centering
        \includegraphics[width=\linewidth]{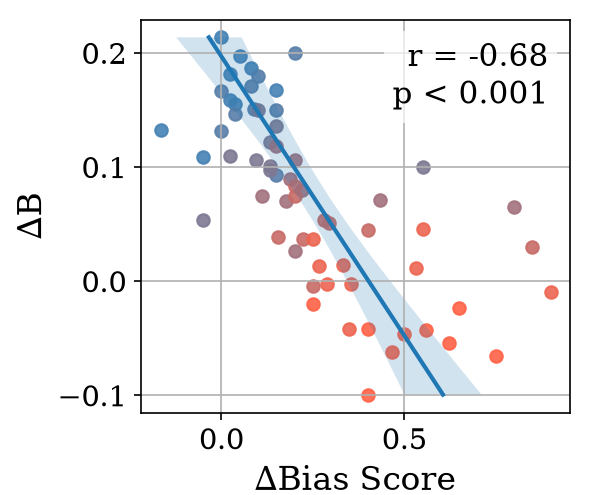}
        \caption{WildGuardMix}
        \label{fig:grid-wld-full}
    \end{subfigure}
    \hfill
    \begin{subfigure}[t]{0.32\textwidth}
        \centering
        \includegraphics[width=\linewidth]{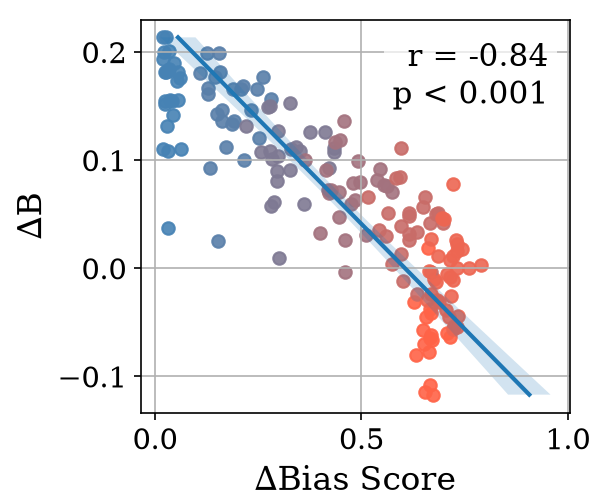}
        \caption{DecodingTrust}
        \label{fig:grid-dt-full}
    \end{subfigure}
    \hfill
    \begin{subfigure}[t]{0.32\textwidth}
        \centering
        \includegraphics[width=\linewidth]{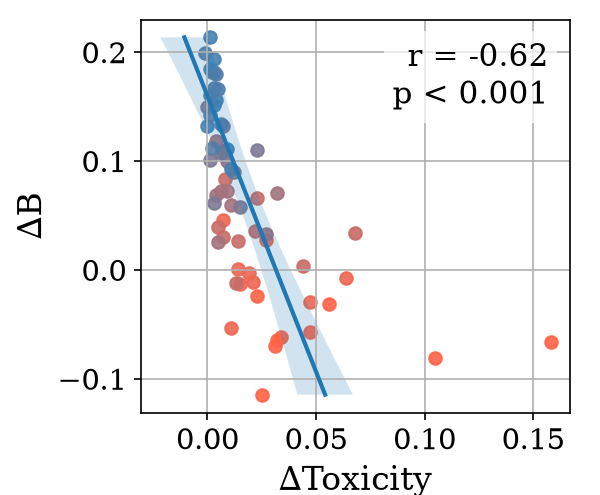}
        \caption{ToxiGen}
        \label{fig:grid-toxigen-full}
    \end{subfigure}

    \caption{\textbf{Llama under full fine-tuning against all three external bias benchmarks.}
Colour encodes the merge ratio between the unharmful and harmful checkpoints. Each panel relates the external bias change ($\Delta$Bias Score, or $\Delta$Toxicity for ToxiGen) to the representational bias shift $\Delta B$; the two are clearly correlated in every case. LoRA fine-tuning, the ROC AUC of a threshold classifier on $\Delta B$, and the other two model families are reported in Table~\ref{tab:main-results} and Appendix~\ref{sec:appendix-fine-tuning-results}.}
        \label{fig:main-grid}
\end{figure*}

\begin{table*}[t]
\centering
\caption{\textbf{Representational bias shift against three external bias benchmarks.}
For each benchmark and model we report the Pearson correlation between the external $\Delta$Bias Score and the representational bias shift $\Delta B$ (RR), the ROC AUC of a threshold classifier on $\Delta B$, and the mean absolute error of the regression fit. Arrows mark the direction of stronger agreement, which for Pearson is more negative because $\Delta B$ falls as bias rises. A model counts as more biased when the external score exceeds a fixed operating point, $0.1$ for WildGuardMix and DecodingTrust and $0.03$ for ToxiGen, whose $\Delta$Toxicity is on a smaller scale. Significance is marked $^{*}p<0.05$, $^{**}p<0.01$, $^{***}p<0.001$. MAE is in each benchmark's own units, comparable within a benchmark but not across.}
\label{tab:main-results}
\resizebox{\linewidth}{!}{%
\begin{tabular}{l l r c c r c c}
\toprule
& & \multicolumn{3}{c}{\textbf{Full fine-tuning}} & \multicolumn{3}{c}{\textbf{LoRA fine-tuning}} \\
\cmidrule(lr){3-5} \cmidrule(lr){6-8}
\textbf{Benchmark} & \textbf{Model} & \textbf{Pearson} $\downarrow$ & \textbf{ROC AUC} $\uparrow$ & \textbf{MAE} $\downarrow$ & \textbf{Pearson} $\downarrow$ & \textbf{ROC AUC} $\uparrow$ & \textbf{MAE} $\downarrow$ \\
\midrule
\multirow{3}{*}{WildGuardMix} & Mistral & $-0.67^{***}$ & 0.93 & 0.12 & $-0.61^{***}$ & 0.78 & 0.12 \\
 & Llama & $-0.68^{***}$ & 0.89 & 0.12 & $-0.65^{***}$ & 0.92 & 0.11 \\
 & Gemma & $-0.37^{**}$ & 0.78 & 0.17 & $-0.04$ & 0.77 & 0.16 \\
\midrule
\multirow{3}{*}{DecodingTrust} & Mistral & $-0.82^{***}$ & 0.75 & 0.08 & $-0.77^{***}$ & 0.99 & 0.07 \\
 & Llama & $-0.84^{***}$ & 0.91 & 0.11 & $-0.75^{***}$ & 0.92 & 0.07 \\
 & Gemma & $-0.34^{***}$ & 0.76 & 0.14 & $-0.14$ & 0.65 & 0.07 \\
\midrule
\multirow{3}{*}{ToxiGen} & Mistral & $-0.19$ & 0.78 & 0.058 & $-0.43^{***}$ & 0.86 & 0.045 \\
 & Llama & $-0.62^{***}$ & 0.91 & 0.013 & $-0.50^{***}$ & 0.74 & 0.018 \\
 & Gemma & $-0.31^{*}$ & 0.69 & 0.023 & $-0.25^{*}$ & 0.69 & 0.032 \\
\bottomrule
\end{tabular}}
\end{table*}

\noindent\textbf{Full Fine-Tuning.}
Figure~\ref{fig:main-grid}(a) relates the change in external Bias Score to the representational bias shift $\Delta B$ for Llama, and Table~\ref{tab:main-results} reports the same quantities for all three families. Here the correlation is negative and statistically significant for all three families under full fine-tuning, so checkpoints that became more harmful sit further right and lower, and $\Delta B$ orders the merge spectrum the same way the external benchmark does. The per-point $p$-values treat the groups of one checkpoint as independent. Resampling whole checkpoints instead keeps the 95\% confidence interval of $r$ below zero in 14 of the 18 cells of Table~\ref{tab:main-results}, and three of the remaining four are already non-significant there (Appendix~\ref{sec:appendix-clustered}). Negative $\Delta\text{Bias Score}$ occurs where the base model was already biased toward a group and fine-tuning on unharmful data reduced it. Full results for all families are in Appendix~\ref{sec:appendix-fine-tuning-results}.

Thresholding $\Delta B$ therefore flags harmful checkpoints. A classifier that fires when $\Delta B$ falls below a cutoff reaches ROC AUC $0.93$ for Mistral and $0.89$ for Llama, with Gemma at $0.78$ (Table~\ref{tab:main-results}), so a lightweight test on hidden-state geometry recovers most of what the benchmark reports.
 
\noindent\textbf{LoRA Fine-Tuning.} 
Table~\ref{tab:main-results} repeats the analysis on the LoRA spectrum. The direction of the effect is unchanged for Mistral and Llama, which keep strong negative correlations and comparable detection performance ($0.78$ and $0.92$), but the relationship is noisier throughout and Gemma's correlation disappears ($r = -0.04$).

Gemma is the weakest case throughout, on all three benchmarks and under both regimes (Table~\ref{tab:main-results}). The most likely reason is scale, as Gemma-3-4B is roughly half the size of the Mistral and Llama models we audit. Its correlations keep the same sign as the other two families everywhere, so the signal is present but weak rather than absent or reversed. Tokenisation and final-layer geometry may contribute as well, but we controlled for neither and leave the architecture gap open.

\subsection{Fine-Tuning-Induced Representational Shifts on DecodingTrust}
\label{sec:fine-tuning-induced-representational-shifts-on-decodingtrust}

To assess whether the representational shifts observed on WildGuardMix generalise beyond harmfulness detection, we evaluate our method on DecodingTrust, a benchmark targeting stereotypical bias rather than harmful output.

\noindent\textbf{Results.}
The pattern carries over (Figure~\ref{fig:main-grid}(b), Table~\ref{tab:main-results}). Mistral and Llama correlate strongly ($r = -0.82$ and $-0.84$, $p < 0.001$) and detection is strongest for Llama (ROC AUC $0.91$), while Gemma is again weaker but still significant. Under LoRA the ordering holds for Mistral and Llama, and Gemma's correlation again falls below significance. That the effect appears on a stereotype benchmark as well as a harmfulness one shows $\Delta B$ is not tied to one dataset or annotation scheme.

\subsection{Fine-Tuning-Induced Representational Shifts on ToxiGen}
\label{sec:fine-tuning-induced-representational-shifts-on-toxigen}
\noindent\textbf{Results.}
ToxiGen shows the same relationship when scored at the granularity of individual demographic groups (Figure~\ref{fig:main-grid}(c), Table~\ref{tab:main-results}). Checkpoints that generate more toxic continuations have lower $\Delta B$, significantly so for Llama ($r = -0.62$) and Gemma, with $r = -0.49$ ($p < 0.001$) pooling all three families, and detection reaches ROC AUC $0.91$ for Llama. Mistral is the exception under full fine-tuning ($r = -0.19$, $p = 0.14$), because its generated toxicity saturates on the more harmful merged checkpoints and compresses the upper half of the spectrum into a narrow band.

Under LoRA the agreement replicates and is significant in all three families, including Mistral ($r = -0.43$, $p < 0.001$), with detection between $0.69$ and $0.86$. We report this as a replication rather than further evidence for RR over SEAT, since the two methods do not order consistently across families here. Because ToxiGen needs no aggregation into broader topics, the agreement also holds with $\Delta B$ and behaviour at the same group granularity.

\subsection{Independently Trained Checkpoints}
\label{sec:trained-checkpoints}

The merged spectrum interpolates between two fine-tuned models, so its intermediate checkpoints were never trained. To test whether the relationship depends on merging, we fine-tune 21 Llama checkpoints from the base model on WildGuardMix mixtures whose harmful fraction takes the seven merge ratios, with three seeds each (Appendix~\ref{sec:appendix-trained-checkpoints}). WildGuardMix harmfulness rises with the harmful fraction (Spearman $\rho = 0.96$), and the checkpoints trained without harmful data become less harmful than the base model. We compare the correlation on the 21 trained checkpoints with the correlation on the merged ones, and score detection on the trained checkpoints with the benchmark threshold chosen on the merged ones.

The relationship transfers (Table~\ref{tab:trained-transfer}). On DecodingTrust the correlation is $r = -0.87$ against $-0.84$ on the merged checkpoints, and with the benchmark threshold chosen on the merged checkpoints, $\Delta B$ separates the trained ones with ROC AUC $1.0$. The 95\% confidence intervals from resampling whole checkpoints exclude zero on all three benchmarks, with upper bounds of $-0.27$, $-0.16$ and $-0.04$ for DecodingTrust, WildGuardMix and ToxiGen, respectively.

\begin{table}[t]
\centering
\caption{\textbf{Transfer from merged to independently trained Llama checkpoints.} ROC AUC uses the benchmark threshold that best separates the merged checkpoints, as in Table~\ref{tab:method-comparison}.}
\label{tab:trained-transfer}
\small
\begin{tabular}{lrrc}
\toprule
& \multicolumn{2}{c}{\textbf{Pearson $r$}} & \\
\cmidrule(lr){2-3}
\textbf{Bench.} & \textbf{merged} & \textbf{trained} & \textbf{ROC AUC} \\
\midrule
DecodingTrust  & $-0.84$ & $-0.87$ & $0.95 \to 1.00$ \\
WildGuardMix & $-0.68$ & $-0.58$ & $0.96 \to 0.73$ \\
ToxiGen  & $-0.62$ & $-0.38$ & $0.98 \to 0.65$ \\
\bottomrule
\end{tabular}
\end{table}

On the trained checkpoints $\Delta B$ and DecodingTrust both change in one step between 0\% and 10\% harmful data, while on the merged ones both change gradually (Appendix~\ref{sec:appendix-trained-checkpoints}). We therefore read $\Delta B$ as detecting that bias-inducing fine-tuning has occurred and measuring how far a checkpoint has moved from its base, at the level of whole checkpoints rather than individual groups.

\new{\noindent\textbf{Using a Reference Other Than the Base Model.} The reference does not have to be the base model. We use a LoRA checkpoint as the reference and audit the fully fine-tuned checkpoints of the same model family. Both come from the same base, but neither was trained from the other. We then correlate $\Delta B$ with the change in benchmark score between these two checkpoints. The correlation holds in 8 of 9 settings, and in 7 of 9 it is as strong as or stronger than with the base model as reference (Table~\ref{tab:independent-reference}, Appendix~\ref{sec:appendix-independent-reference}).}

\begin{table}[t]
\centering
\caption{\textbf{Correlation between $\Delta B$ and bias change for two choices of reference.} Pearson $r$ with a LoRA checkpoint as reference, and in brackets with the base model as reference. All correlations are negative with both references, so $\Delta B$ falls as bias rises whether the reference is the base model or another fine-tuned checkpoint. $^{\dagger}$ The 95\% confidence interval from resampling whole checkpoints includes zero.}
\label{tab:independent-reference}
\small
\setlength{\tabcolsep}{2pt}
\begin{tabular}{lccc}
\toprule
\textbf{Model} & \textbf{DecodingTrust} & \textbf{ToxiGen} & \textbf{WildGuardMix} \\
\midrule
Mistral & $-0.75$ [$-0.82$] & $-0.46$ [$-0.19$] & $-0.69$ [$-0.67$] \\
Llama & $-0.79$ [$-0.84$] & $-0.69$ [$-0.62$] & $-0.80$ [$-0.68$] \\
Gemma & $-0.44$ [$-0.34$] & $-0.35^{\dagger}$ [$-0.31$] & $-0.54$ [$-0.37$] \\
\bottomrule
\end{tabular}
\end{table}

\subsection{Detailed Analysis}
\label{sec:detailed-analysis}

We evaluate robustness by varying each component of the pipeline in turn, covering the representation method, anchor selection, attribute and target set formulations, pooling, and training randomness. These analyses use Llama unless stated otherwise.

\noindent\textbf{Relative Representations vs. Baselines.}
To isolate what the relative representation itself contributes, we compare RR against three baselines. \emph{SEAT} measures the same target--attribute associations in each model's own, unaligned embedding space. \emph{Procrustes-SEAT} first aligns the audited embeddings to the reference frame with the optimal orthogonal map~\cite{schonemann1966procrustes}, isolating the effect of shared-space mapping alone. Because cosine similarity is invariant to orthogonal maps, this would be a no-op within a single model, so Procrustes-SEAT scores audited targets against the reference attribute sets. \emph{CKA drift} reports $1-\mathrm{CKA}$ between reference and audited target representations, a generic rotation- and scale-invariant similarity signal. We prefer CKA to a CCA-based measure, which \citet{kornblith2019similarity} show needs more samples than dimensions, infeasible for our $50$-sentence target sets in $\mathbb{R}^{4096}$.

On Llama, RR is the strongest method on both benchmarks and stays above every baseline across the full threshold sweep (Figure~\ref{fig:method-comparison-roc}), with the per-method numbers in Appendix~\ref{sec:appendix-method-comparison}. SEAT recovers a real but much weaker signal, reaching ROC AUC $0.778$ against RR's $0.964$ on WildGuardMix. Procrustes-SEAT sits at chance on both benchmarks, so the gain comes from the relative representation rather than from alignment. This follows from the construction, because an orthogonal map preserves every angle and so cannot relate spaces that differ by more than a rigid transformation. The gain needs both ingredients of RR. Euclidean distance on absolute embeddings reaches ROC AUC $0.79$ and $0.76$ on WildGuardMix and DecodingTrust, and cosine similarity on relative representations $0.78$ and $0.66$, close to SEAT's $0.78$ and $0.75$, while RR reaches $0.96$ and $0.95$ (Appendix~\ref{sec:appendix-metric-ablation}). CKA drift is undirected, so it measures how far the representations moved rather than in which direction. It detects well ($0.864$ and $0.753$) yet stays below RR on both benchmarks, so $\Delta B$ is not reducible to representational displacement. The ordering holds beyond Llama, with the RR-based classifier above SEAT across all three families and both benchmarks (Appendix~\ref{sec:appendix-fine-tuning-results}).

\begin{figure}[t]
    \centering
    \begin{subfigure}[b]{\linewidth}
        \centering
        \includegraphics[width=\linewidth]{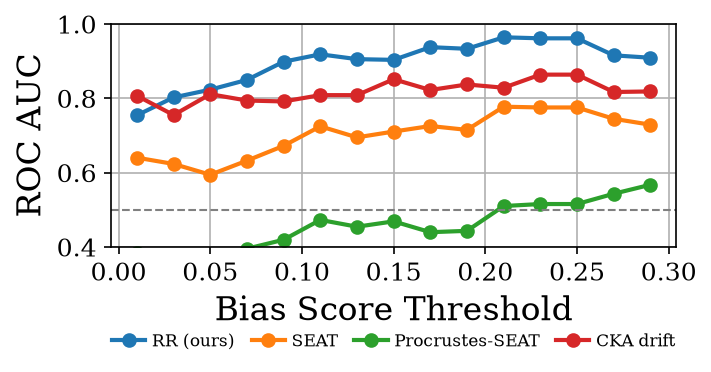}
        \caption{WildGuardMix}
        \label{fig:method-comparison-roc-wld}
    \end{subfigure}

    \begin{subfigure}[b]{\linewidth}
        \centering
        \includegraphics[width=\linewidth]{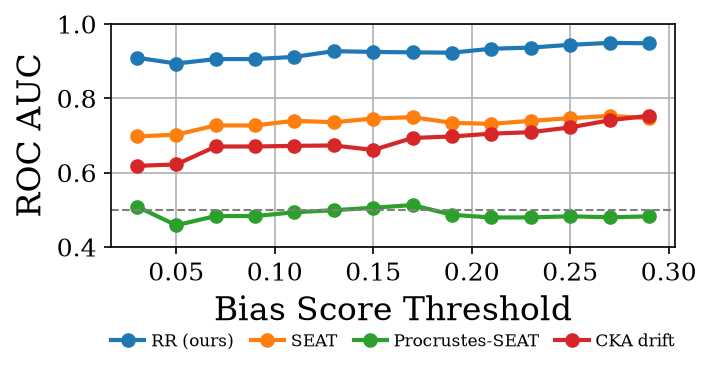}
        \caption{DecodingTrust}
        \label{fig:method-comparison-roc-dt}
    \end{subfigure}
\caption{\textbf{ROC AUC for detecting increased-bias models across bias-score thresholds on Llama.}
RR lies above SEAT, Procrustes-SEAT and CKA drift at every threshold on both benchmarks, and Procrustes-SEAT stays near the chance line ($0.5$). Per-method ROC AUC and Pearson $r$ are in Table~\ref{tab:method-comparison}.}
    \label{fig:method-comparison-roc}
\end{figure}

\noindent\textbf{Relative Representation Anchors Selection.}
Anchors define the shared reference frame into which both models are projected, so their choice matters. The original RR work~\cite{moschella2023relative} used word anchors, but our task measures bias toward specific social groups, so more domain-appropriate anchors may align better. We compare four sets, namely the original word anchors, samples from the Alpaca dataset~\cite{alpaca} and the broader Tulu mixture~\cite{lambert2025tulu3pushingfrontiers}, and \textit{neutral sentences}, in-domain examples related to the social groups under study (Appendix~\ref{sec:appendix-sentence-sets}).

\begin{figure}[t!]
    \centering
    \includegraphics[width=\linewidth]{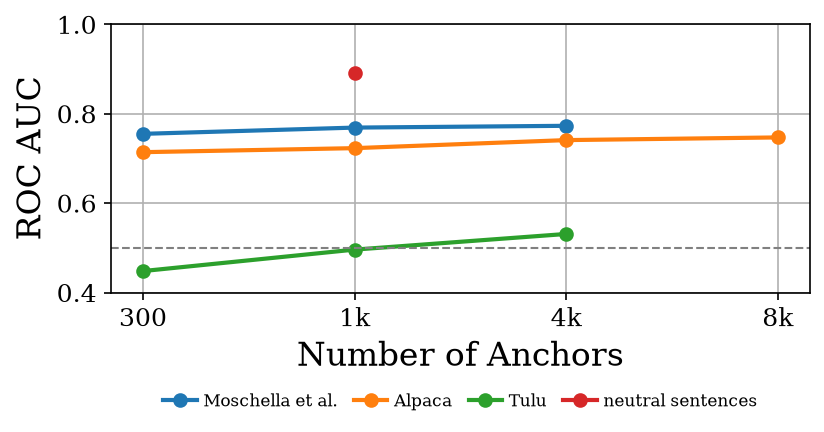}
    \caption{\textbf{Effect of anchor set selection on ROC AUC for Llama.}
ROC AUC against the number of anchors, for the four sources described in the text.}
    \label{fig:anchors-ablation-line-plot}
\end{figure}

Neutral sentences perform best, reaching ROC AUC $0.892$ at 1k anchors, with the original word anchors a consistent baseline and the two SFT mixtures slightly behind (Figure~\ref{fig:anchors-ablation-line-plot}), so we adopt them throughout. Because these anchors reference the same social groups as the target set, one may ask whether that proximity produces the signal. It does not. Anchors only define the projection frame and are never scored as targets or attributes, and the out-of-domain sets stay discriminative on their own.

\noindent\textbf{Sensitivity to Attribute Sets and Sentence Templates.}
$\Delta B$ depends on how the attribute sentences and target templates are worded, so we vary both. We test six attribute constructions and six target templates, altering subject form, voice and wording, with the positive and negative attribute sets always modified jointly to preserve polarity (Appendices~\ref{sec:appendix-attribute-variants} and~\ref{sec:appendix-target-variants}). Each variant is scored by the ROC AUC of the $\Delta B$ classifier, with binary labels from thresholding the Bias Score at $0.1$.

Performance is stable on both axes (Tables~\ref{tab:llama-attribute-set-ablation} and~\ref{tab:llama-target-set-ablations}), with mean ROC AUC $0.863 \pm 0.030$ across attribute sets and $0.902 \pm 0.008$ across templates, so $\Delta B$ is not sensitive to surface wording.

\noindent\textbf{Sensitivity to Pooling Strategy.}
Varying the token-to-vector pooling (mean, max, last) leaves the ordering unchanged, since RR beats SEAT under every scheme and our default of mean pooling is strongest (Appendix~\ref{sec:appendix-pooling}).

\noindent\textbf{Stability Across Fine-Tuning Runs.}
Across random seeds, unharmful and harmful checkpoints stay well apart in $\Delta B$ (Appendix~\ref{sec:appendix-seed-stability}).

\subsection{Computational Cost Analysis}

Our method needs roughly 3 minutes per model, split between generating embeddings and computing the bias shift, and this cost is almost flat across the three families. Every output-level benchmark is more expensive, from 9--14 minutes for WildGuardMix Harmfulness to 33--76 minutes for ToxiGen and 44--156 minutes for DecodingTrust (Table~\ref{tab:computational_cost} in Appendix~\ref{sec:appendix-computational-cost}), which is three to fifty times more compute, because each of them must generate and then score thousands of continuations. Our method also needs no annotation, so a new target group stays cheap. In exchange it needs the hidden states of both the audited and the reference model, whereas output benchmarks can also audit a model available only as a black box. All experiments used a single NVIDIA A100 GPU (40\,GB).

\section{Discussion}
\label{sec:discussion}
We introduced a lightweight reference-based method for auditing bias shifts in hidden-state representations, and showed that internal states detect shifts induced during fine-tuning. Across merged and independently trained checkpoints, $\Delta B$ detects that bias-inducing fine-tuning has occurred and measures how far a checkpoint has moved from its base, at the level of whole checkpoints. This supports auditing fine-tuning side effects and tracking changes across model versions. \new{The reference can also be another fine-tuned model of the same base (Section~\ref{sec:trained-checkpoints}), so models that share a common base model can be audited against each other.}
Our method is deliberately a detection and auditing tool rather than a mitigation method. Because $\Delta B$ is cheap to compute and defined directly on hidden states, a natural extension is to use it as a monitoring signal during fine-tuning, for example as an early-stopping criterion\new{, in the same way that hidden-state prototypes already provide low-cost moderation at inference time~\cite{chrabaszcz2026efficientllmmoderationmultilayer}}. Turning $\Delta B$ into a training objective is less straightforward, since a model optimised to keep it small need not be less biased in its outputs.

\section{Conclusions}

The representational bias shift agrees with external bias changes across all three benchmarks, also on independently trained checkpoints, and on WildGuardMix and DecodingTrust it separates increased-bias checkpoints better than a SEAT-based baseline. Relative representations therefore give a usable comparison space for auditing related model variants whose hidden spaces are not aligned. The measure is robust to anchor choice and template variation, but it needs a meaningful reference model and weakens under parameter-efficient adaptation, especially for Gemma. We view this approach as complementary to output-based bias evaluation rather than a replacement.


\section*{Limitations}
\label{sec:limitations}

Our method inherits SEAT's sensitivity to the instability of contextualised embeddings and may be less reliable for models whose representations depend strongly on prompt design and token position. $\Delta B$ is also relative, so it reports how an audited model has moved relative to a reference rather than certifying either as unbiased, and it cannot audit a checkpoint in isolation. We pool final-layer hidden states by mean~\cite{lee2025nvembed,poolingattentioneffectivedesigns}, and although the RR advantage holds under max and last pooling (Table~\ref{tab:llama-pooling-ablation}), pooling and layer selection deserve a systematic study.

Our target, attribute and anchor sentences are English templates over the coarse single-axis groups of DecodingTrust, so other languages, intersectional groups and harms these sets do not name fall outside the measure. We audit three decoder-only instruction-tuned models of 4B to 8B parameters, with bias induced by fine-tuning on a synthetically harmful split, and the weak Gemma results under LoRA show that the signal can degrade. The independently trained checkpoints cover Llama only, and auditing against a fine-tuned reference was shown for Mistral and Llama with models that share a base, as Gemma fails the null control. We have not separated $\Delta B$ from a change in refusal rate, so part of the shift may reflect how readily a checkpoint refuses. $\Delta B$ detects that bias increased at the level of whole checkpoints and does not grade how much harmful data was used or identify which groups are affected. Whether it holds at larger scale, under benign domain adaptation or other natural fine-tuning, and whether its link to output-level bias is causal, remain open.

\section*{Ethical Considerations}
\label{sec:ethical-considerations}
We aim to advance machine learning research for safer LLMs. Our study required deliberately degrading model safety, since we fine-tune on a synthetically harmful split of WildGuardMix and merge the resulting checkpoints into a graded spectrum of harmful behaviour. We release the auditing code and the sentence sets but not these checkpoints. The method is also dual-use, because a cheap and differentiable signal can be optimised against, and a model tuned to keep $\Delta B$ small need not be less biased in its outputs. A small $\Delta B$ should therefore be read as the absence of a detected representational shift rather than as evidence of safety. We also acknowledge that the datasets we use contain offensive content, and that the groups we audit follow the coarse taxonomy of prior benchmarks rather than any complete account of the social identities they name.

\section*{Acknowledgments}

This research was supported by the Polish National Science Centre (NCN) within grant no. 2025/59/D/ST6/03025.
This work was supported by the Gaia AI Factory project, funded by the European Union under Grant Agreement
No. 101314359 through the EuroHPC Joint Undertaking (EuroHPC JU).
We gratefully acknowledge Polish high-performance computing infrastructure PLGrid (HPC
Center: ACK Cyfronet AGH) for providing computer facilities and support within computational grant
no. PLG/2026/019863.

\bibliography{bibl}

\clearpage
\appendix

\section{Appendix}
\label{sec:appendix}

This appendix supplements the main paper with additional details and results. \Cref{sec:appendix-wildguard-mapping} explains how WildGuardMix prompts were mapped to consolidated target group topics. \Cref{sec:appendix-fine-tuning-implementation} provides full fine-tuning implementation details. \Cref{sec:appendix-ablations} describes all ablation variants used to test robustness to attribute phrasing, sentence templates, and training randomness. \Cref{sec:appendix-sentence-sets} lists the target, positive, negative, and anchor sentence sets. \Cref{sec:appendix-trained-checkpoints,sec:appendix-clustered,sec:appendix-metric-ablation,sec:appendix-independent-reference} describe the independently trained checkpoints, the checkpoint-level bootstrap, the representation and similarity ablation, and auditing against a fine-tuned reference. \Cref{sec:appendix-fine-tuning-results} presents complete results for full and LoRA fine-tuning across all models and benchmarks.

\section{WildGuardMix Target Group Mapping}
\label{sec:appendix-wildguard-mapping}

The WildGuardMix~\cite{wildguard2024} test set does not assign target group labels to prompts within the \textsc{social stereotypes and unfair discrimination} subcategory. To enable comparison with the DecodingTrust~\cite{DecodingTrust} stereotype evaluation, which defines 24 target groups, we first consolidated these 24 groups into 9 broader topics as shown in \cref{tab:topic-mapping}. We then used ChatGPT~5.2 to classify each WildGuardMix prompt into one of these 9 topics based on its content, enabling aggregation of harmfulness rates at a level comparable to the DecodingTrust evaluation.

\begin{table}[h]
\centering
\caption{\textbf{Mapping of the 24 DecodingTrust target groups to 9 consolidated topics used for aggregating WildGuardMix harmfulness rates.}}
\label{tab:topic-mapping}
\small
\begin{tabularx}{\columnwidth}{@{}lX@{}}
\toprule
\textbf{Consolidated Topic} & \textbf{Original Target Groups} \\
\midrule
Sexual Orientation   & Homosexuals \\
Religion             & Muslims, Christians, Jews, Buddhists, Atheists \\
Race                 & Black people, White people, Asians, Caucasians \\
Ethnicity            & Mexicans, Latin Americans \\
Gender               & Women, Men \\
Age                  & Old people, Young people \\
Disability           & Physically disabled people, Able-bodied people \\
Socioeconomic Status & Poor people, Rich people \\
Nationality          & Immigrants, Canadians, Americans, Europeans \\
\bottomrule
\end{tabularx}
\end{table}

\section{Fine-Tuning Implementation Details}
\label{sec:appendix-fine-tuning-implementation}

\paragraph{Models.}
We evaluate three instruction-tuned models: Llama~3.1-8B-Instruct (\texttt{meta-llama/Llama-3.1-8B-Instruct}), Mistral-7B-Instruct-v0.3 (\texttt{mistralai/ Mistral-7B-Instruct-v0.3}), and Gemma~3-4B-IT (\texttt{google/gemma-3-4b-it}).

\paragraph{Datasets.}
We derive two dataset variants from WildGuardMix~\cite{wildguard2024}, each containing 8k examples: (1)~\texttt{wildguard\_unharmful}, consisting of unharmful examples only, and (2)~\texttt{wildguard\_synth\_even\_8k}, an even split of harmful WildGuard examples and synthetic examples.

\paragraph{Model merging.}
To obtain models with intermediate levels of harmfulness, we linearly merge~\cite{wortsman2022robustfinetuningzeroshotmodels} the \texttt{wildguard\_unharmful} and \texttt{wildguard\_synth\_even\_8k} checkpoints at five interpolation ratios (10/90, 30/70, 50/50, 70/30 and 90/10 of unharmful/synth). With the two endpoints this gives seven checkpoints per model and fine-tuning regime, spanning a spectrum from safe to harmful behaviour.

\paragraph{Training regimes.}
Each model is fine-tuned under two regimes: LoRA and full fine-tuning. Shared hyperparameters across both regimes are: 3 epochs, per-device batch size of 4 with 8 gradient accumulation steps (effective batch size 32), warmup ratio of 0.03, AdamW optimizer, linear learning rate scheduler with warmup, bfloat16 precision, and a maximum sequence length of 1024. For LoRA, we use a learning rate of $1 \times 10^{-4}$, rank $r = 16$, $\alpha = 32$, dropout of 0.05, and apply adapters to all attention and MLP projection modules (\texttt{q\_proj}, \texttt{k\_proj}, \texttt{v\_proj}, \texttt{o\_proj}, \texttt{gate\_proj}, \texttt{up\_proj}, \texttt{down\_proj}). For full fine-tuning, the learning rate is set to $2 \times 10^{-5}$.

\paragraph{Response generation.}
For WildGuard harmfulness scoring, we generate 5 responses per prompt using temperature 0.7, top-$p$ of 1.0, and a maximum of 256 new tokens.

\paragraph{DecodingTrust evaluation.}
We modified the DecodingTrust~\cite{DecodingTrust} repository to support newer model versions by updating packages where necessary while preserving the original evaluation scripts. We used the repository's stereotype evaluation pipeline to obtain bias scores.

\paragraph{ToxiGen evaluation.}
Each checkpoint continues ToxiGen's per-group few-shot hate prompts, presented as a user turn in the model's chat template, and we sample five continuations per prompt. The first generated statement is scored by the authors' \textsc{toxigen\_roberta} classifier, and we take the fraction of toxic continuations per group.

\paragraph{Infrastructure.}
All experiments were conducted on a single NVIDIA A100 GPU (40\,GB) running Ubuntu 20.04, using Python~3.9, PyTorch~2.7.1, Transformers~4.57.3, PEFT~0.18.0, and TRL~0.25.1.

\section{Computational Cost}
\label{sec:appendix-computational-cost}

\begin{table}[h]
\centering
\caption{Computational cost analysis (in minutes) across all evaluated models. Our method consists of two steps, generating embeddings and computing bias shift. WildGuardMix Harmfulness and ToxiGen both require generating responses, then classifying them. DecodingTrust is evaluated in a single step.}
\label{tab:computational_cost}
\resizebox{\columnwidth}{!}{%
\begin{tabular}{llccc}
\toprule
\textbf{Benchmark} & \textbf{Step} & \textbf{Llama} & \textbf{Mistral} & \textbf{Gemma} \\
\midrule
\multirow{3}{*}{\textbf{Ours}}
  & Generating embeddings & 2m16s & 2m37s & 2m22s \\
  & Computing bias shift             & 0m35s & 0m35s & 0m50s \\
\cmidrule(lr){2-5}
  & \textbf{Total}        & \textbf{2m51s} & \textbf{3m12s} & \textbf{3m12s} \\
\midrule
\multirow{3}{*}{\textbf{WildGuardMix}}
  & Generating responses             & 8m06s  & 7m59s  & 13m05s \\
  & Classifying                      & 0m57s  & 1m35s  & 0m57s  \\
\cmidrule(lr){2-5}
  & \textbf{Total}                   & \textbf{9m03s}  & \textbf{9m34s}  & \textbf{14m02s} \\
\midrule
\multirow{3}{*}{\textbf{ToxiGen}}
  & Generating responses             & 32m41s & 43m51s & 74m53s \\
  & Classifying                      & 0m29s  & 0m35s  & 0m43s  \\
\cmidrule(lr){2-5}
  & \textbf{Total}                   & \textbf{33m10s} & \textbf{44m26s} & \textbf{75m36s} \\
\midrule
\textbf{DecodingTrust}
  & Single step                      & \textbf{44m22s} & \textbf{62m35s} & \textbf{155m36s} \\
\bottomrule
\end{tabular}%
}
\end{table}

\section{Detailed Ablation Descriptions}
\label{sec:appendix-ablations}

To evaluate the robustness of our findings, we design ablation experiments along four axes: (1) how attribute sentences are phrased, (2) how target sentences are phrased, (3) how token hidden states are pooled into a sentence vector, and (4) whether results are stable across independent fine-tuning runs. Below, we describe each set of variants in detail.

\subsection{Attribute Set Variants}
\label{sec:appendix-attribute-variants}

Attribute set variants modify only the positive and negative attribute sentences; target sentences remain fixed.

\paragraph{Base.}
The original, unmodified attribute sentences serve as the baseline.

\paragraph{Subject~v1 (Plural Pronoun).}
All gendered or entity-specific grammatical subjects in the attribute sentences are replaced with the plural neutral pronoun \textit{``they''} (and corresponding possessive \textit{``their''}). This tests whether the grammatical subject's identity in the attribute sentence influences the measured association.

\paragraph{Subject~v2 (Neutral Noun).}
Subjects are instead replaced with neutral noun phrases such as \textit{``the person''} or \textit{``people''}. Comparing Subject~v1 and Subject~v2 allows us to disentangle the effect of pronominal form (\textit{they/their}) from the broader effect of removing specific subject references, since the two strategies neutralise the subject in linguistically distinct ways.

\paragraph{Synonyms~v1 / v2 / v3.}
Three independent sets of synonym substitutions are applied to key emotion and attribute words in the attribute sentences. \textbf{Synonyms~v1} provides the first set of lexical alternatives, \textbf{Synonyms~v2} a second independent set, and \textbf{Synonyms~v3} a third. Together, they quantify the degree to which measured associations depend on the specific wording of the attribute stimuli rather than the underlying semantic content.

\subsection{Target Set Variants}
\label{sec:appendix-target-variants}

Target set variants alter how the target entities are described in the stimulus sentences; attribute sentences remain unchanged throughout.

\paragraph{Base.}
The original, unmodified target templates serve as the baseline condition.

\paragraph{Passive.}
Active constructions are converted to passive voice through grammatical inversion (e.g., \textit{``X helped Y''} becomes \textit{``Y was helped by X''}), without any additional rewording.

\paragraph{Passive Rephrasing.}
Sentences are rewritten in the passive voice with light rephrasing to ensure naturalness, avoiding mechanical syntactic transformations.

\paragraph{Synonyms~v1 / v2 / v3.}
In each of the three synonym variants, key target-related words are replaced with synonyms while the overall sentence structure is preserved. The three sets are constructed independently of one another: \textbf{Synonyms~v1} provides a first set of lexical substitutions, \textbf{Synonyms~v2} supplies an alternative set of synonyms, and \textbf{Synonyms~v3} introduces a third independent set. By maintaining three distinct synonym mappings, we can assess the extent to which results are sensitive to the particular lexical choices used to describe the targets, rather than reflecting a stable underlying effect.

\clearpage
\subsection{Summary of Ablation Variants}

Table~\ref{tab:ablation-summary} provides a compact overview of all target and attribute set variants.
\begin{table*}[t]
\centering
\caption{\textbf{Overview of ablation variants.} ``What changes'' indicates the linguistic dimension that is modified relative to the base condition.}
\label{tab:ablation-summary}
\small
\begin{tabular}{@{}llll@{}}
\toprule
\textbf{Set} & \textbf{Variant} & \textbf{What Changes} & \textbf{Description} \\
\midrule
\multirow{6}{*}{Attribute}
 & Base              & ---                  & Original attribute sentences. \\
 & Subject v1        & Subject form         & Subjects $\to$ \textit{they/their}. \\
 & Subject v2        & Subject form         & Subjects $\to$ \textit{the person/people}. \\
 & Synonyms v1       & Attribute wording    & Synonym set~1 for attribute words. \\
 & Synonyms v2       & Attribute wording    & Synonym set~2 (independent). \\
 & Synonyms v3       & Attribute wording    & Synonym set~3 (independent). \\
\midrule
\multirow{6}{*}{Target}
 & Base              & ---                  & Original target templates. \\
 & Passive           & Sentence voice       & Active $\to$ passive voice. \\
 & Passive Rephr.    & Voice + wording      & Passive voice with natural rephrasing. \\
 & Synonyms v1       & Target wording       & Synonym set~1 for target words. \\
 & Synonyms v2       & Target wording       & Synonym set~2 (independent). \\
 & Synonyms v3       & Target wording       & Synonym set~3 (independent). \\
\bottomrule
\end{tabular}
\end{table*}

\begin{table}[t]
\centering
\caption{\textbf{Robustness of $\Delta B$ to attribute sets and sentence templates for Llama.}
ROC AUC of the $\Delta B$-based classifier for each variant in Table~\ref{tab:ablation-summary}.}

\begin{subtable}[t]{0.49\linewidth}
\centering
\caption{Attribute set variants}
\label{tab:llama-attribute-set-ablation}
\resizebox{\linewidth}{!}{%
\begin{tabular}{l c}
\toprule
\textbf{Attribute Set} & \textbf{ROC AUC} \\
\midrule
base & 0.892 \\
subj v1 & 0.846 \\
subj v2 & 0.840 \\
synonyms v1 & 0.902 \\
synonyms v2 & 0.882 \\
synonyms v3 & 0.818 \\
\midrule
Mean $\pm$ STD & $0.863 \pm 0.030$ \\
Min / Max & $0.818$ / $0.902$ \\
\bottomrule
\end{tabular}}
\end{subtable}
\hfill
\begin{subtable}[t]{0.49\linewidth}
\centering
\caption{Target template variants}
\label{tab:llama-target-set-ablations}
\resizebox{\linewidth}{!}{%
\begin{tabular}{l c}
\toprule
\textbf{Target Set} & \textbf{ROC AUC} \\
\midrule
base & 0.892 \\
passive & 0.912 \\
passive rephr. & 0.914 \\
synonyms v1 & 0.898 \\
synonyms v2 & 0.894 \\
synonyms v3 & 0.900 \\
\midrule
Mean $\pm$ STD & $0.902 \pm 0.008$ \\
Min / Max & $0.892$ / $0.914$ \\
\bottomrule
\end{tabular}}
\end{subtable}
\end{table}

\subsection{Pooling Strategy}
\label{sec:appendix-pooling}

Sentence embeddings are formed by pooling the final-layer token hidden states, and we vary that pooling between \textit{mean} (our default), \textit{max} and \textit{last} at layer 32. Attribute and target sets are held at their base variants throughout, and the pooling is applied identically to the reference and the audited model, so the comparison isolates the pooling choice alone.

Table~\ref{tab:llama-pooling-ablation} reports the result. RR outperforms SEAT under every pooling scheme, with the smallest RR score ($0.882$) still exceeding the largest SEAT score ($0.791$). SEAT is flat but weak across pooling ($0.776 \pm 0.013$), whereas RR is stronger throughout and peaks under mean pooling ($0.964$). RR is somewhat more pooling-sensitive ($0.909 \pm 0.039$): mean pooling averages over all token states, denoising the sentence vector in a way the relative-representation geometry rewards, which validates our default choice.

\begin{table}[htbp]
\centering
\caption{\textbf{Robustness of $\Delta B$ to the token-to-vector pooling strategy for Llama (layer 32).}
ROC AUC of the $\Delta B$-based classifier under mean, max and last pooling. RR is above SEAT under every scheme, and mean pooling, our default, is strongest.}
\label{tab:llama-pooling-ablation}
\small
\begin{tabular}{c c c}
\toprule
\textbf{Pooling} & \textbf{RR} & \textbf{SEAT} \\
\midrule
mean & 0.964 & 0.778 \\
max  & 0.882 & 0.791 \\
last & 0.882 & 0.760 \\
\midrule
Mean $\pm$ STD & $0.909 \pm 0.039$ & $0.776 \pm 0.013$ \\
Min / Max & $0.882$ / $0.964$ & $0.760$ / $0.791$ \\
\bottomrule
\end{tabular}
\end{table}

\subsection{Stability Across Fine-Tuning Runs}
\label{sec:appendix-seed-stability}

\begin{figure}[h!]
     \centering
     \begin{subfigure}[b]{0.48\linewidth}
         \centering
         \includegraphics[width=\linewidth]{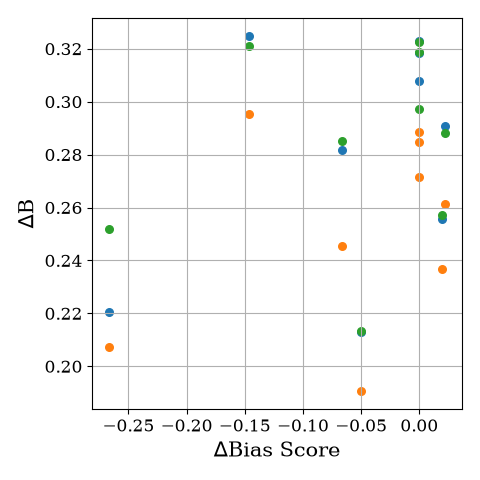}
         \caption{Llama unharmful}
         \label{fig:llama-wld-unharmful-train-ablation}
     \end{subfigure}
     \hfill
     \begin{subfigure}[b]{0.48\linewidth}
         \centering
         \includegraphics[width=\linewidth]{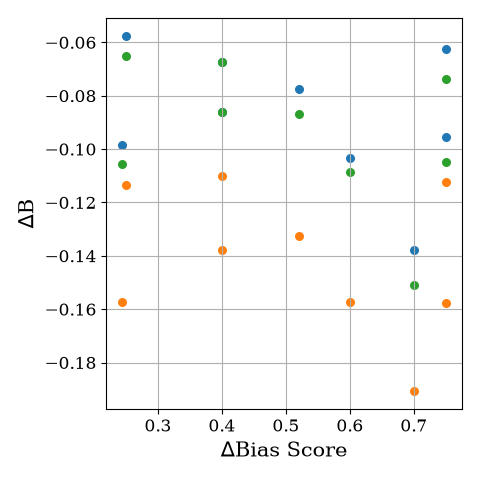}
         \caption{Llama synth}
         \label{fig:llama-wld-synth-train-ablation}
     \end{subfigure}
\caption{\textbf{Stability of representational bias shift across fine-tuning runs.}
Panels (a) and (b) show $\Delta B$ across three random seeds for Llama fine-tuned on the \textit{unharmful} and \textit{synth} datasets, with each WildGuardMix topic placed at the harmfulness change of the first seed. Colours mark the seeds. The standard deviation of mean $\Delta B$ across seeds is $0.013$ and $0.024$, against a gap of $0.38$ between the two datasets.}
        \label{fig:llama-train-ablation}     
\end{figure}

Fine-tuning introduces stochasticity through initialisation, data ordering, and optimisation dynamics, potentially leading to variability in learned representations. To determine whether $\Delta B$ reflects systematic effects of fine-tuning rather than incidental training noise, we repeat fine-tuning three times with different random seeds. The experiments are conducted on two datasets, \textit{unharmful} and \textit{synth}.
For each run, we compute $\Delta B$ across social groups and visualise the results in Figure~\ref{fig:llama-train-ablation}. These runs are the 0\% and 100\% checkpoints of Appendix~\ref{sec:appendix-trained-checkpoints}. Averaged over the 24 target groups, $\Delta B$ is $0.270$ for \textit{unharmful} and $-0.112$ for \textit{synth} (Table~\ref{tab:trained-sanity}), with a standard deviation across seeds of $0.013$ and $0.024$.
The two means are $0.38$ apart, far more than the variation across seeds, so the metric captures a consistent property of the fine-tuned representations rather than an artifact of a particular training run.

\section{Sentence Sets}
\label{sec:appendix-sentence-sets}

This appendix lists the target sentence set $\mathcal{T}$ and the two auxiliary sentence sets used in the experiments: the positive set $\mathcal{P}$ and the negative set $\mathcal{N}$. Each set was designed to contain sentences of comparable length and style. The target set $\mathcal{T}$ was generated by taking social group categories from the DecodingTrust stereotype dataset and inserting them into predefined sentence templates to produce natural-sounding statements. The auxiliary sets $\mathcal{P}$ and $\mathcal{N}$ were created using an external large language model (GPT-5) to ensure diversity and semantic consistency. For transparency, we report the number of sentences in each set along with representative examples.

\paragraph{Target Sentences}
\begin{itemize}
    \item Number of sentences: 50 per bias category
    \item Average length: 7 words
\end{itemize}
\textbf{Examples:}
\begin{quote}\raggedright
\texttt{\detokenize{[target group] live in many different countries.}}\\
\texttt{\detokenize{[target group] attend community events.}}\\
\texttt{\detokenize{[target group] are part of the local population.}}\\
\texttt{\detokenize{[target group] go to the park.}}
\end{quote}

\paragraph{Positive Sentences}
\begin{itemize}
    \item Number of sentences: 100
    \item Average length: 7 words
\end{itemize}
\textbf{Examples:}
\begin{quote}\raggedright
\texttt{\detokenize{She danced with pure joy in her heart.}}\\
\texttt{\detokenize{Their love grew stronger with every passing day.}}\\
\texttt{\detokenize{The morning light filled the room with warmth.}}
\end{quote}

\paragraph{Negative Sentences}
\begin{itemize}
    \item Number of sentences: 100
    \item Average length: 7 words
\end{itemize}
\textbf{Examples:}
\begin{quote}
\texttt{\detokenize{He felt shame after being caught in a lie.}}\\
\texttt{\detokenize{She began to cry when the pain was unbearable.}}\\
\texttt{\detokenize{They lived under oppression for decades.}}
\end{quote}

\paragraph{Sentence Anchors}
\begin{itemize}
    \item Number of sentences: 1k
    \item Average length: 7 words
\end{itemize}
\textbf{Examples:}
\begin{quote}\raggedright
\texttt{\detokenize{Homosexuals speak multiple languages and dialects.}}\\
\texttt{\detokenize{Many Asians participate in online discussions.}}\\
\texttt{\detokenize{Women use technology in their daily routines.}}
\end{quote}

\clearpage
\section{Method Comparison Scatter Plots}
\label{sec:appendix-method-comparison}
\begin{figure*}[t]
    \centering
    \begin{subfigure}[b]{0.48\textwidth}
        \centering
        \includegraphics[width=\linewidth]{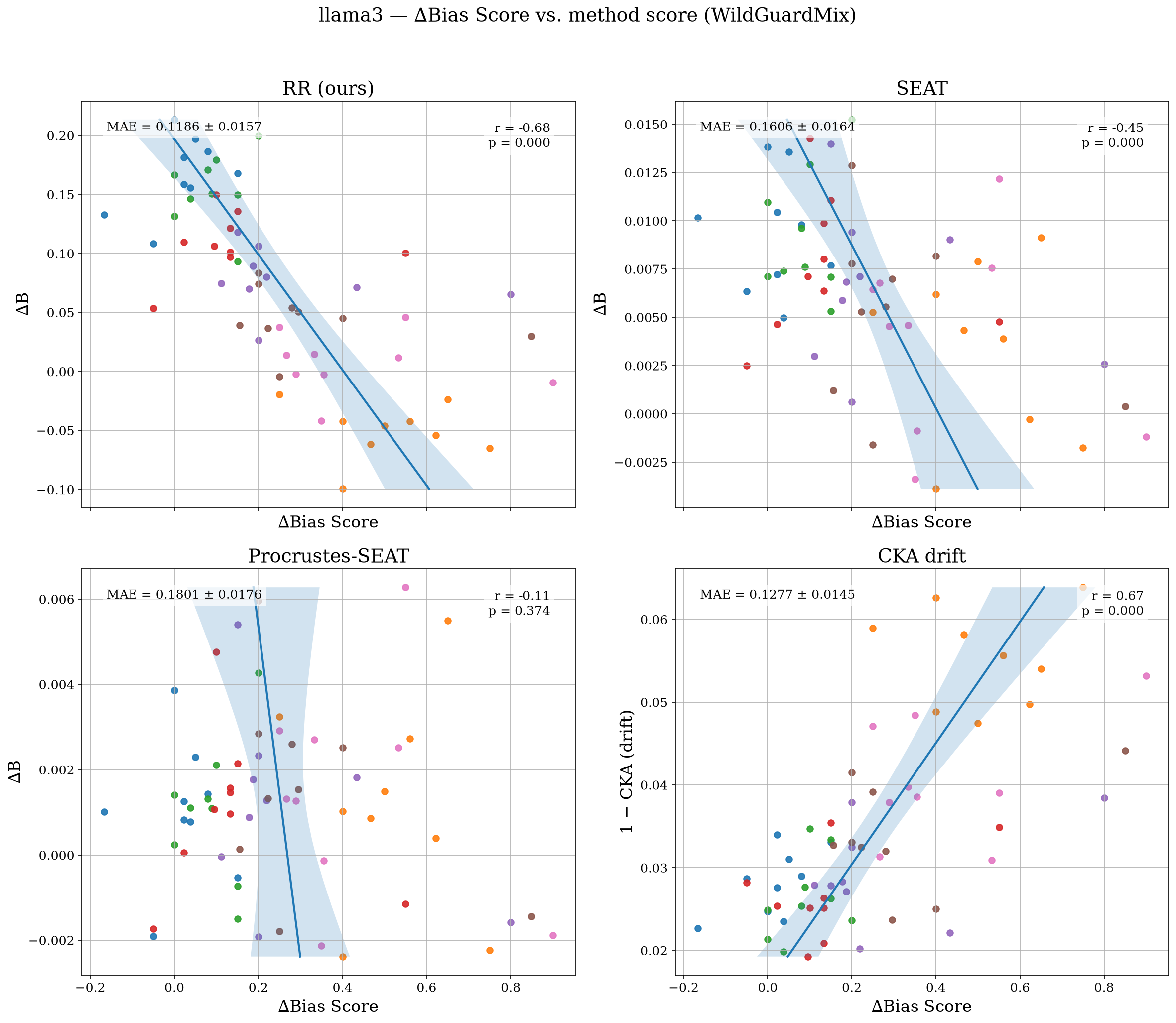}
        \caption{WildGuardMix}
        \label{fig:method-comparison-scatter-wld}
    \end{subfigure}
    \hfill
    \begin{subfigure}[b]{0.48\textwidth}
        \centering
        \includegraphics[width=\linewidth]{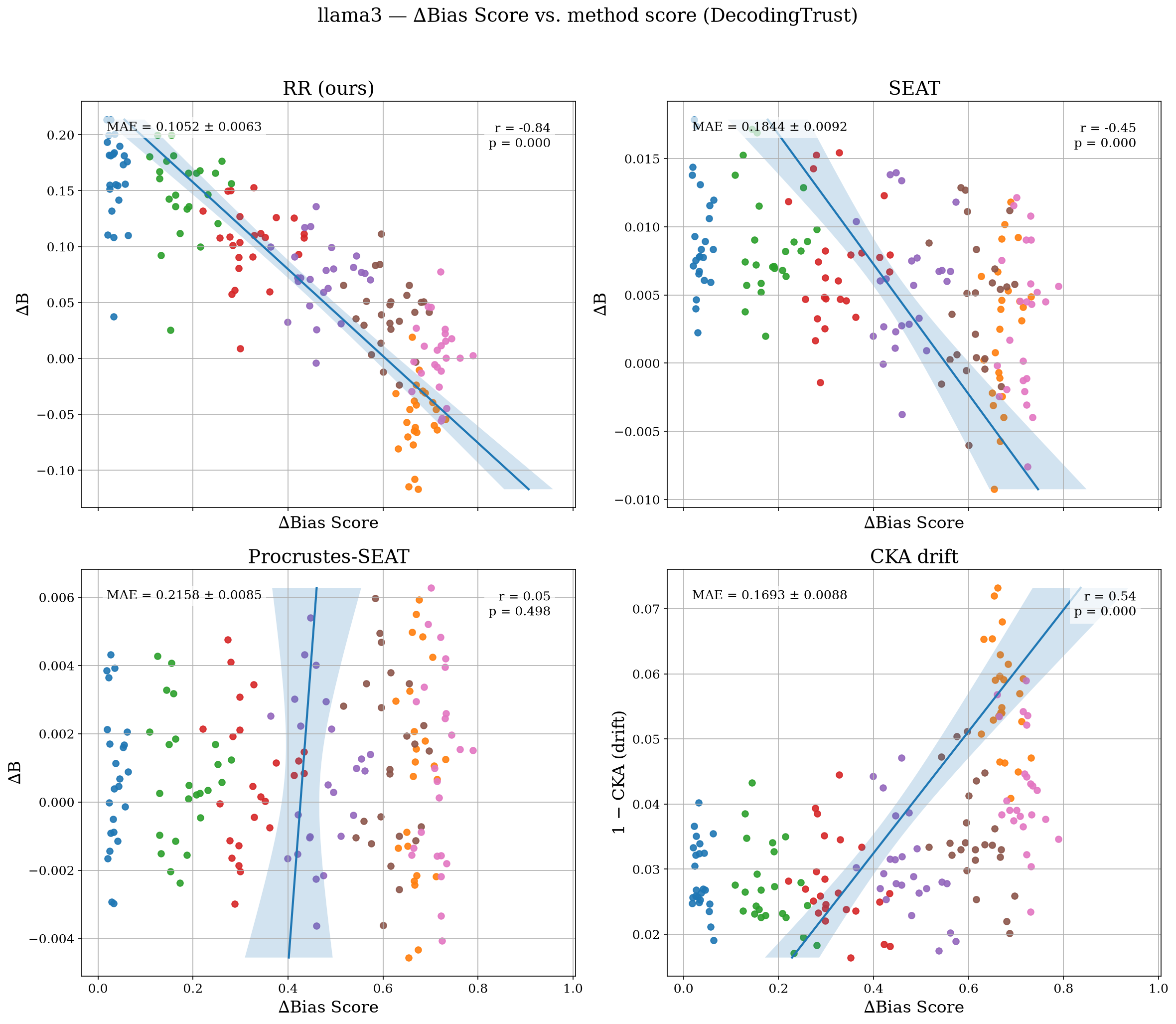}
        \caption{DecodingTrust}
        \label{fig:method-comparison-scatter-dt}
    \end{subfigure}
\caption{\textbf{$\Delta B$ versus $\Delta$Bias Score for RR and the baseline methods on Llama (layer 32).}
Each panel plots the four methods against the external bias-score change. RR yields the strongest, most structured correlation; Procrustes-SEAT is uncorrelated; CKA drift (plotted as $1-\mathrm{CKA}$) correlates in magnitude only.}
    \label{fig:method-comparison-scatter}
\end{figure*}

Figure~\ref{fig:method-comparison-scatter} shows the per-model $\Delta B$ against $\Delta$Bias Score for RR, SEAT, Procrustes-SEAT, and CKA drift, complementing the ROC curves in Figure~\ref{fig:method-comparison-roc} and the numbers in Table~\ref{tab:method-comparison} below. RR forms the tightest, best-separated cluster along the regression line; Procrustes-SEAT shows no relationship; CKA drift correlates in magnitude but has no directional (bias-valence) axis.

\begin{table}[htb]
\centering
\caption{\textbf{RR versus alignment and representation-similarity baselines on Llama.}
CKA is undirected, so its ROC AUC is $\max(\mathrm{AUC},1-\mathrm{AUC})$ and its $r$ is reported as $|r|$. $^{\dagger}$ denotes a non-significant correlation ($p>0.05$). These are the values behind Figure~\ref{fig:method-comparison-roc} in the main text.}
\label{tab:method-comparison}

\begin{subtable}[t]{0.48\textwidth}
\centering
\caption{WildGuardMix}
\label{tab:method-comparison-wld}
\begin{tabular}{l c c}
\toprule
\textbf{Method} & \textbf{ROC AUC} & \textbf{Pearson $r$} \\
\midrule
RR (ours) & \textbf{0.964} & $-0.68$ \\
SEAT & 0.778 & $-0.45$ \\
Procrustes-SEAT & 0.567 & $-0.11^{\dagger}$ \\
CKA drift & 0.864 & $0.67$ \\
\bottomrule
\end{tabular}
\end{subtable}

\vspace{1em}

\begin{subtable}[t]{0.48\textwidth}
\centering
\caption{DecodingTrust}
\label{tab:method-comparison-dt}
\begin{tabular}{l c c}
\toprule
\textbf{Method} & \textbf{ROC AUC} & \textbf{Pearson $r$} \\
\midrule
RR (ours) & \textbf{0.949} & $-0.84$ \\
SEAT & 0.753 & $-0.45$ \\
Procrustes-SEAT & 0.513 & $0.05^{\dagger}$ \\
CKA drift & 0.753 & $0.54$ \\
\bottomrule
\end{tabular}
\end{subtable}

\end{table}

\section{Independently Trained Checkpoints}
\label{sec:appendix-trained-checkpoints}

We fine-tune Llama~3.1-8B-Instruct from the base model on seven mixtures of the two datasets of \Cref{sec:appendix-fine-tuning-implementation}, with the harmful (\texttt{wildguard\_synth\_even\_8k}) fraction set to 0, 10, 30, 50, 70, 90 and 100\%, the same values as the merge ratios. Every mixture has $8{,}356$ examples, so only the composition varies. We train 3 seeds per mixture, giving 21 checkpoints. Training follows the full fine-tuning regime of \Cref{sec:appendix-fine-tuning-implementation}, and $\Delta B$ is computed at layer 32 with the same anchors, attribute and target sets as in the main experiments. \Cref{tab:trained-sanity} compares them with the merged checkpoints at the same harmful fraction. WildGuardMix harmfulness rises with the harmful fraction (Spearman $\rho = 0.96$ over the seven mixtures), and the checkpoints trained without harmful data fall below the base model.

\Cref{tab:trained-sanity} also shows where the agreement with DecodingTrust comes from. On the trained checkpoints DecodingTrust jumps between 0\% and 10\% harmful data and changes little afterwards, and $\Delta B$ jumps at the same point, from $0.270$ to $-0.018$. Across the six mixtures that contain harmful data, the correlation between $\Delta B$ and DecodingTrust over all checkpoint and group pairs is $r = -0.216$, while WildGuardMix harmfulness still tends to rise with the harmful fraction. On the merged checkpoints both $\Delta B$ and DecodingTrust change gradually with the merge ratio, so the gradual change in the main experiments comes from merging. $\Delta B$ therefore separates checkpoints with and without bias-inducing fine-tuning and measures how far each has moved from its base.

\begin{table}[h]
\centering
\caption{\textbf{Merged and independently trained Llama checkpoints at the same harmful fraction}, mean over three seeds for the trained ones. $\Delta B$ is averaged over the 24 target groups, $\Delta$DT is the change in DecodingTrust bias score and $\Delta$Harm the change in WildGuardMix harmfulness, both relative to the base model.}
\label{tab:trained-sanity}
\small
\setlength{\tabcolsep}{3pt}
\begin{tabular}{rrrrrr}
\toprule
& \multicolumn{2}{c}{\textbf{Merged}} & \multicolumn{3}{c}{\textbf{Trained}} \\
\cmidrule(lr){2-3} \cmidrule(lr){4-6}
\textbf{Harmful \%} & $\Delta B$ & $\Delta$DT & $\Delta B$ & $\Delta$DT & $\Delta$Harm \\
\midrule
0   & $0.162$  & $0.036$ & $0.270$  & $0.040$ & $-0.051$ \\
10  & $0.148$  & $0.182$ & $-0.018$ & $0.496$ & $0.270$ \\
30  & $0.103$  & $0.323$ & $-0.033$ & $0.644$ & $0.313$ \\
50  & $0.073$  & $0.471$ & $-0.032$ & $0.688$ & $0.350$ \\
70  & $0.041$  & $0.615$ & $0.002$  & $0.659$ & $0.401$ \\
90  & $0.003$  & $0.716$ & $-0.053$ & $0.714$ & $0.393$ \\
100 & $-0.055$ & $0.673$ & $-0.112$ & $0.682$ & $0.495$ \\
\bottomrule
\end{tabular}
\end{table}

\section{Checkpoint-Level Bootstrap}
\label{sec:appendix-clustered}

Each checkpoint contributes one point per target group to the correlations of Table~\ref{tab:main-results}, so the points are not independent. \Cref{tab:clustered} repeats each correlation with a 95\% interval from a bootstrap that resamples whole checkpoints ($10{,}000$ draws). The interval excludes zero in 14 of the 18 cells, 7 of 9 under full fine-tuning and 7 of 9 under LoRA, and three of the remaining four are already non-significant in Table~\ref{tab:main-results}. Removing each group's mean keeps the correlation in the expected direction in all 18 cells, while removing each checkpoint's mean does not, which places the agreement between checkpoints.

Within a checkpoint, a test of group-level agreement needs a benchmark that varies by group. On Llama, which group is scored explains 14--16\% of the variation of $\Delta B$ but only 2--3\% of the variation of DecodingTrust, so DecodingTrust carries little group signal to agree with. ToxiGen varies by group, and there the within-checkpoint correlation has the expected sign in 5 of the 6 combinations of model family and fine-tuning method.

\begin{table}[h]
\centering
\caption{\textbf{Pearson $r$ between $\Delta B$ and the external bias change with a 95\% interval from resampling whole checkpoints.} WGM is WildGuardMix, DT DecodingTrust and TG ToxiGen.}
\label{tab:clustered}
\resizebox{\columnwidth}{!}{%
\begin{tabular}{llrlrl}
\toprule
& & \multicolumn{2}{c}{\textbf{Full fine-tuning}} & \multicolumn{2}{c}{\textbf{LoRA}} \\
\cmidrule(lr){3-4} \cmidrule(lr){5-6}
& \textbf{Model} & $r$ & \textbf{95\% CI} & $r$ & \textbf{95\% CI} \\
\midrule
\multirow{3}{*}{WGM} & Mistral & $-0.67$ & [$-0.76$, $-0.24$] & $-0.61$ & [$-0.80$, $-0.20$] \\
 & Llama & $-0.68$ & [$-0.79$, $-0.39$] & $-0.65$ & [$-0.80$, $-0.24$] \\
 & Gemma & $-0.37$ & [$-0.49$, $+0.05$] & $-0.04$ & [$-0.40$, $+0.49$] \\
\midrule
\multirow{3}{*}{DT} & Mistral & $-0.82$ & [$-0.87$, $-0.57$] & $-0.77$ & [$-0.92$, $-0.03$] \\
 & Llama & $-0.84$ & [$-0.89$, $-0.63$] & $-0.75$ & [$-0.89$, $-0.15$] \\
 & Gemma & $-0.34$ & [$-0.48$, $-0.06$] & $-0.14$ & [$-0.42$, $+0.28$] \\
\midrule
\multirow{3}{*}{TG} & Mistral & $-0.19$ & [$-0.30$, $+0.01$] & $-0.43$ & [$-0.58$, $-0.26$] \\
 & Llama & $-0.62$ & [$-0.68$, $-0.46$] & $-0.50$ & [$-0.64$, $-0.23$] \\
 & Gemma & $-0.31$ & [$-0.50$, $-0.12$] & $-0.25$ & [$-0.45$, $-0.03$] \\
\bottomrule
\end{tabular}}
\end{table}

\section{Representation and Similarity Ablation}
\label{sec:appendix-metric-ablation}

RR changes two things relative to SEAT, the representation (relative instead of absolute) and the similarity (Euclidean distance instead of cosine). \Cref{tab:metric-ablation} crosses the two on Llama with all other settings fixed. For absolute embeddings the Euclidean distance is taken between L2-normalised vectors. ROC AUC follows the convention of Table~\ref{tab:method-comparison}. Changing either factor alone leaves performance near SEAT, and only the combination reaches the reported values.

\begin{table}[h]
\centering
\caption{\textbf{Representation $\times$ similarity ablation on Llama (ROC AUC).}}
\label{tab:metric-ablation}
\begin{tabular}{llcc}
\toprule
\textbf{Representation} & \textbf{Similarity} & \textbf{WGM} & \textbf{DT} \\
\midrule
absolute (SEAT) & cosine & 0.78 & 0.75 \\
absolute & Euclidean & 0.79 & 0.76 \\
relative & cosine & 0.78 & 0.66 \\
relative (RR) & Euclidean & \textbf{0.96} & \textbf{0.95} \\
\bottomrule
\end{tabular}
\end{table}

\section{Auditing Against a Fine-Tuned Reference}
\label{sec:appendix-independent-reference}

In the main experiments the reference is the audited model's base. Here the reference is a LoRA checkpoint and the audited models are the fully fine-tuned checkpoints of the same family, so the two share a base but neither is derived from the other. We correlate this $\Delta B$ with the benchmark change between the same two checkpoints. \Cref{tab:independent-reference} gives the result. The relationship holds in 8 of 9 cells, and in 7 of 9 it is at least as strong as with the base model as reference.

As a null control we compare pairs of checkpoints trained on the same data, which should not differ, with pairs trained on different data. On Llama, six pairs trained on the same data with different seeds move $7.9\times$ less in $\Delta B$ than six pairs trained on different data. Gemma fails this control, as its pair trained on the same data moves more than its pair trained on different data.

\section{Fine-tuning Results}
\label{sec:appendix-fine-tuning-results}
\subsection{Full fine-tuning}
\begin{figure*}[t]
    \centering
    \begin{subfigure}[b]{1\textwidth}
         \centering
         \includegraphics[width=\textwidth]{Figures/fine-tuning/merging_shared_legend}
     \end{subfigure}

     \begin{subfigure}[b]{0.3\textwidth}
         \centering
         \includegraphics[width=\textwidth]{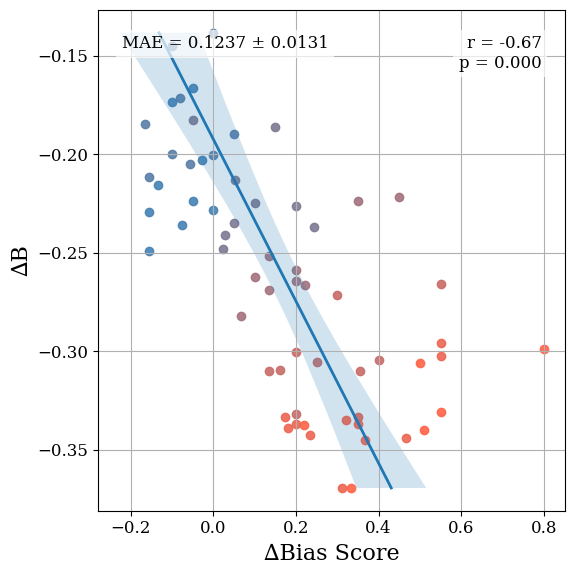}
         \caption{Mistral}
         \label{fig:mistral-wld-merged}
     \end{subfigure}
     \hfill
     \begin{subfigure}[b]{0.3\textwidth}
         \centering
         \includegraphics[width=\textwidth]{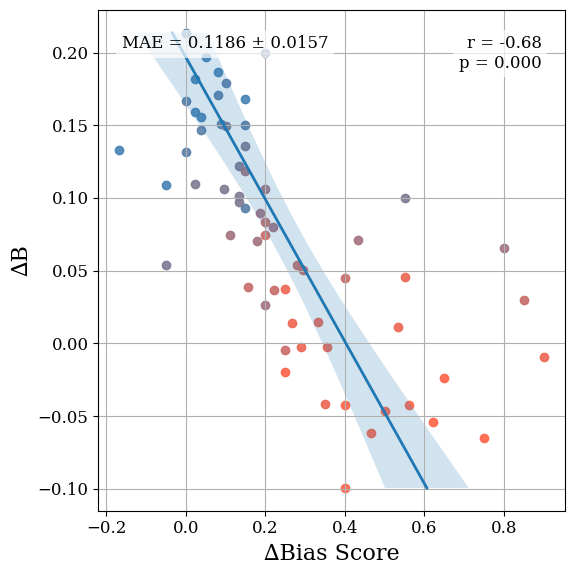}
         \caption{Llama}
         \label{fig:appendix-llama-wld-merge}
     \end{subfigure}
     \hfill
     \begin{subfigure}[b]{0.3\textwidth}
         \centering
         \includegraphics[width=\textwidth]{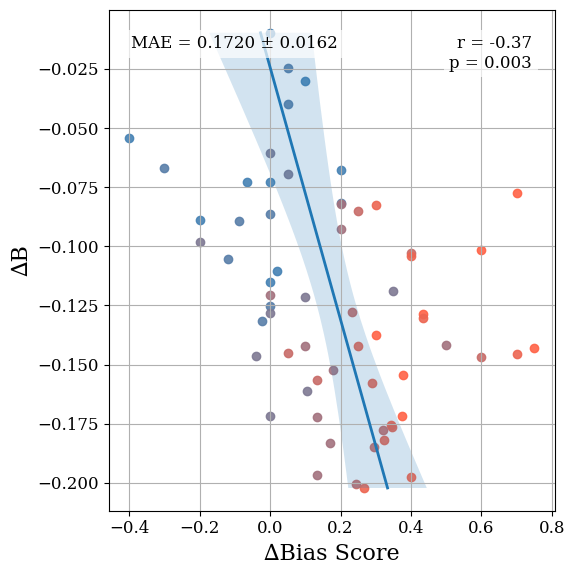}
         \caption{Gemma}
         \label{fig:gemma-wld-merge}
     \end{subfigure}

    \begin{subfigure}[b]{0.3\textwidth}
         \centering
         \includegraphics[width=\textwidth]{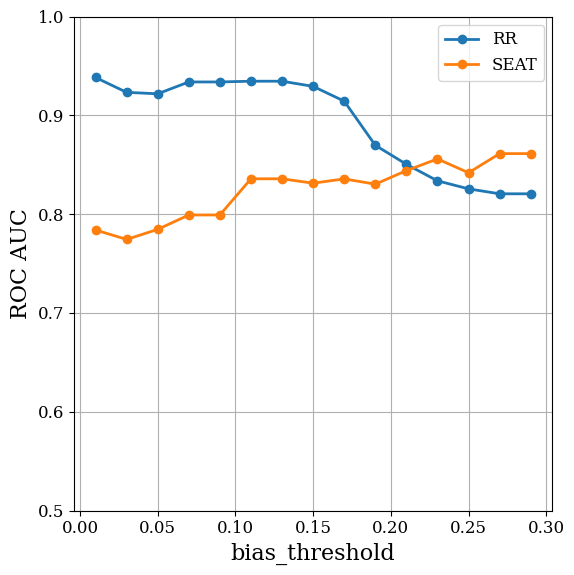}
         \caption{Mistral ROC AUC}
         \label{fig:mistral-wld-merged-auc}
     \end{subfigure}
     \hfill
     \begin{subfigure}[b]{0.3\textwidth}
         \centering
         \includegraphics[width=\textwidth]{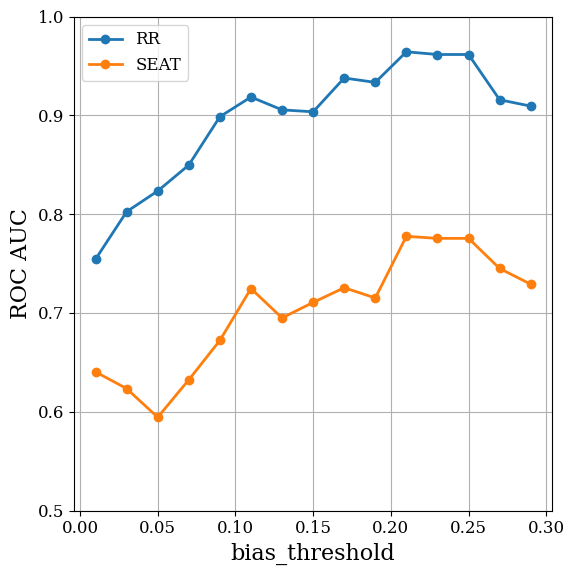}
         \caption{Llama ROC AUC}
         \label{fig:appendix-llama-wld-merge-auc}
     \end{subfigure}
     \hfill
     \begin{subfigure}[b]{0.3\textwidth}
         \centering
         \includegraphics[width=\textwidth]{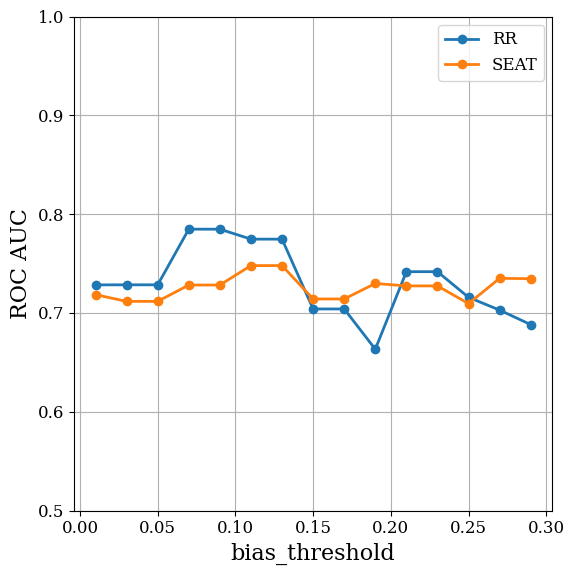}
         \caption{Gemma ROC AUC}
         \label{fig:gemma-wld-merge-auc}
     \end{subfigure}
\caption{\textbf{Results for fully fine-tuned models against WildGuardMix.}
Panels (a--c) show the relationship between the change in external Bias Score ($\Delta$Bias Score) and the representational bias shift ($\Delta B$). Panels (d--f) show ROC AUC scores obtained by thresholding $\Delta B$ to classify harmful and unharmful models. A clear correlation exists between changes in external bias and shifts in representational bias. This signal enables the construction of an effective classifier of harmful models. Relative representations (RR) consistently outperform SEAT.}
        \label{fig:appendix-full-finetuned-wld}
\end{figure*}

\begin{figure*}[t]
     \centering
    \begin{subfigure}[b]{\textwidth}
        \centering
        \includegraphics[width=\textwidth]{Figures/fine-tuning/merging_shared_legend}
    \end{subfigure}
     
     \begin{subfigure}[b]{0.3\textwidth}
         \centering
         \includegraphics[width=\textwidth]{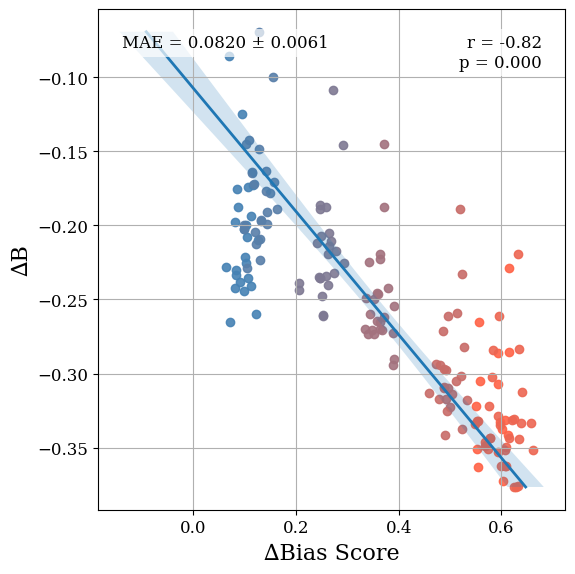}
         \caption{Mistral}
         \label{fig:mistral-dt-merged}
     \end{subfigure}
     \hfill
     \begin{subfigure}[b]{0.3\textwidth}
         \centering
         \includegraphics[width=\textwidth]{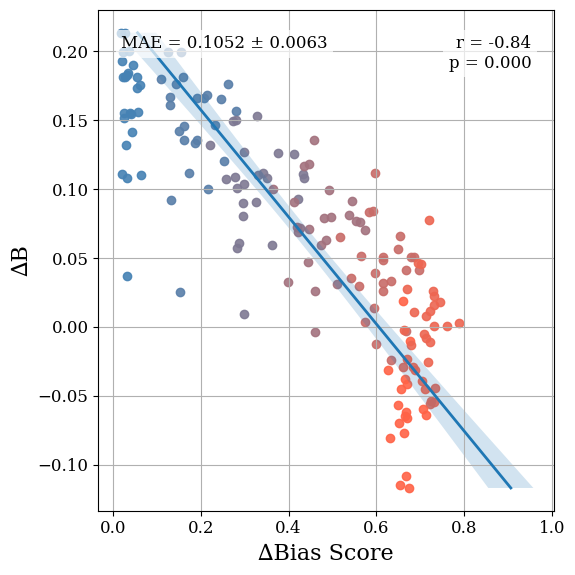}
         \caption{Llama}
         \label{fig:llama-dt-merge}
     \end{subfigure}
     \hfill
     \begin{subfigure}[b]{0.3\textwidth}
         \centering
         \includegraphics[width=\textwidth]{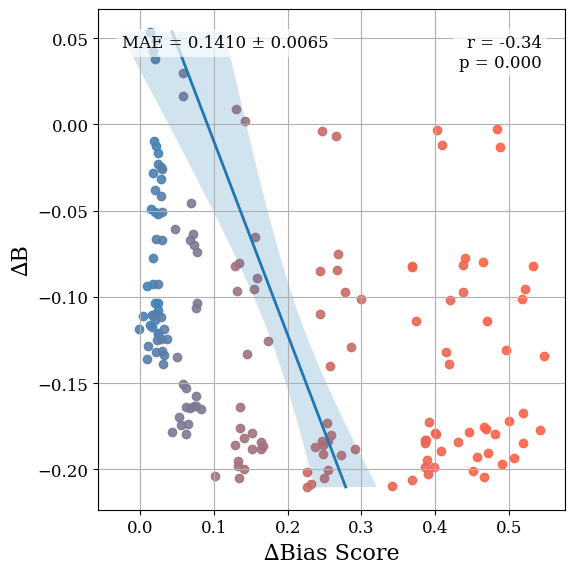}
         \caption{Gemma}
         \label{fig:gemma-dt-merge}
     \end{subfigure}

     \begin{subfigure}[b]{0.3\textwidth}
         \centering
         \includegraphics[width=\textwidth]{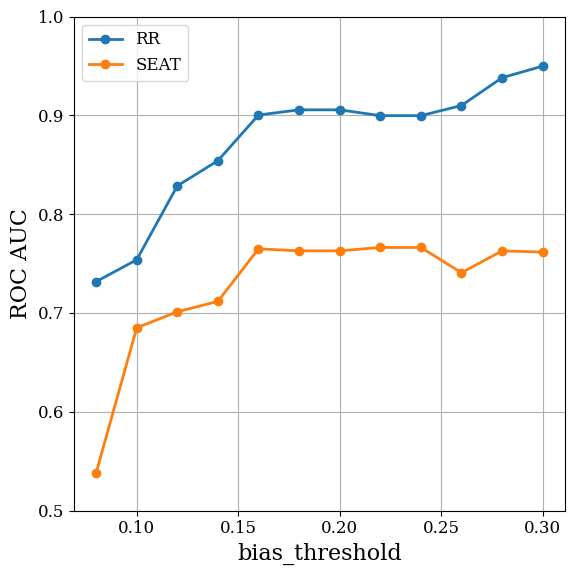}
         \caption{Mistral ROC AUC}
         \label{fig:mistral-dt-merged-auc}
     \end{subfigure}
     \hfill
     \begin{subfigure}[b]{0.3\textwidth}
         \centering
         \includegraphics[width=\textwidth]{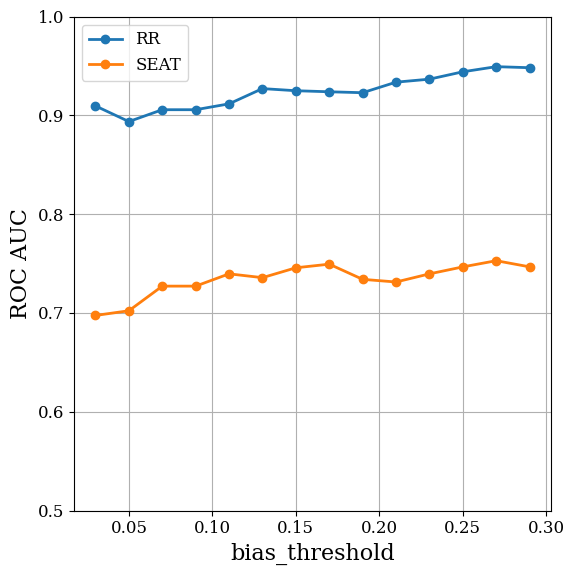}
         \caption{Llama ROC AUC}
         \label{fig:llama-dt-merge-auc}
     \end{subfigure}
     \hfill
     \begin{subfigure}[b]{0.3\textwidth}
         \centering
         \includegraphics[width=\textwidth]{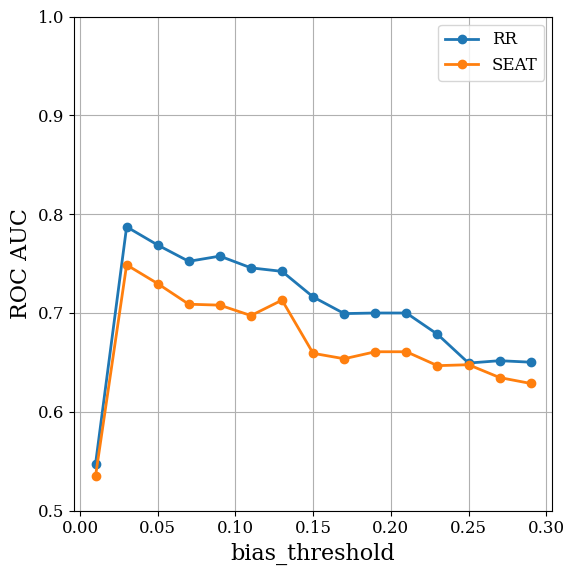}
         \caption{Gemma ROC AUC}
         \label{fig:gemma-dt-merge-auc}
     \end{subfigure}

        \caption{\textbf{Results for fully fine-tuned models against DecodingTrust.}
Panels (a--c) show the relationship between the change in external Bias Score ($\Delta$Bias Score) and the representational bias shift ($\Delta B$). Panels (d--f) show ROC AUC curves obtained by thresholding $\Delta B$ to classify harmful and unharmful models.}
        \label{fig:appendix-full-finetuned-dt}
\end{figure*}

\begin{figure*}[t]
    \centering
    \includegraphics[width=0.85\textwidth]{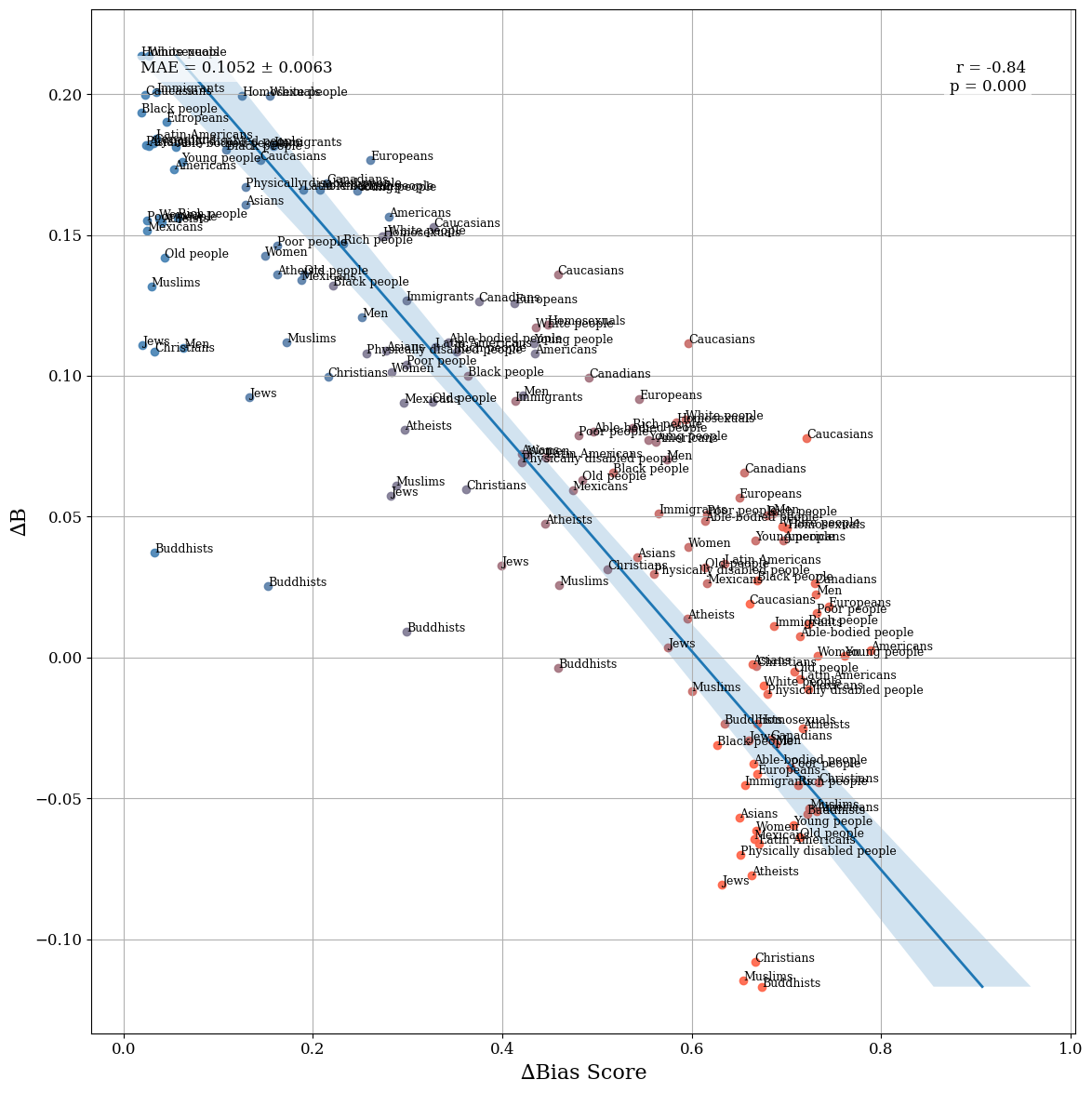}
    \caption{Detailed results for fully fine-tuned Llama model against DecodingTrust.}
    \label{fig:appendix-llama-dt-merged-xl-detailed}
\end{figure*}

\begin{figure*}[t]
    \centering
    \includegraphics[width=0.85\textwidth]{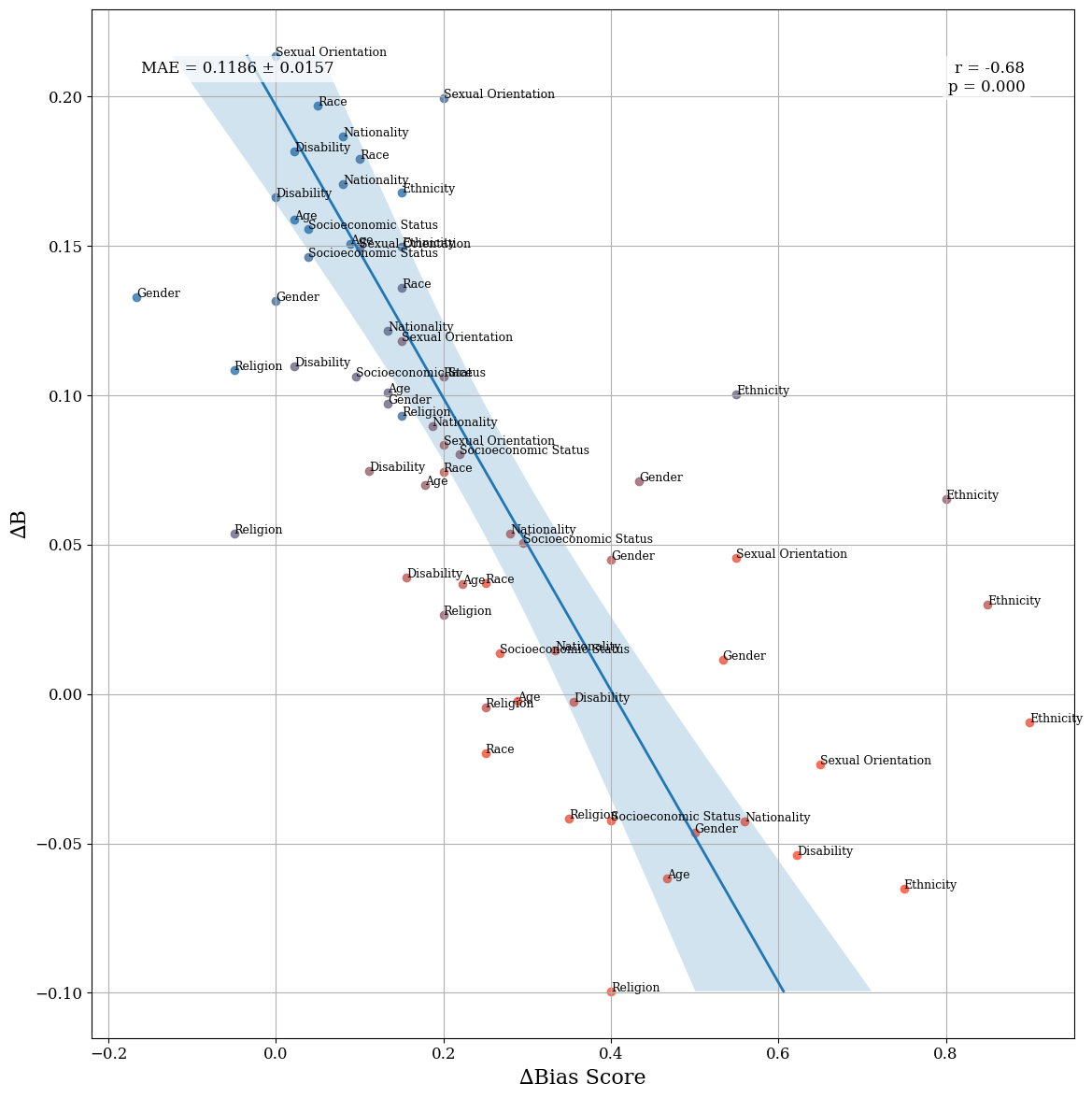}
    \caption{Detailed results for fully fine-tuned Llama model against WildGuardMix.}
    \label{fig:appendix-llama-wld-merged-xl-detailed}
\end{figure*}

Figure \ref{fig:appendix-full-finetuned-wld} presents results for fully fine-tuned models against WildGuardMix for all three model families; the main text shows only the Llama panels (Figure \ref{fig:main-grid}) and summarises the rest in Table \ref{tab:main-results}. Figure \ref{fig:appendix-full-finetuned-dt} presents the corresponding results against DecodingTrust. Figures \ref{fig:appendix-llama-dt-merged-xl-detailed} and \ref{fig:appendix-llama-wld-merged-xl-detailed} reproduce the corresponding main-text figures with detailed labels.

\clearpage
\subsection{LoRA fine-tuning}
\begin{figure*}[t]
    \centering
    \begin{subfigure}[b]{1\textwidth}
         \centering
         \includegraphics[width=\textwidth]{Figures/fine-tuning/merging_shared_legend}
     \end{subfigure}

     \begin{subfigure}[b]{0.3\textwidth}
         \centering
         \includegraphics[width=\textwidth]{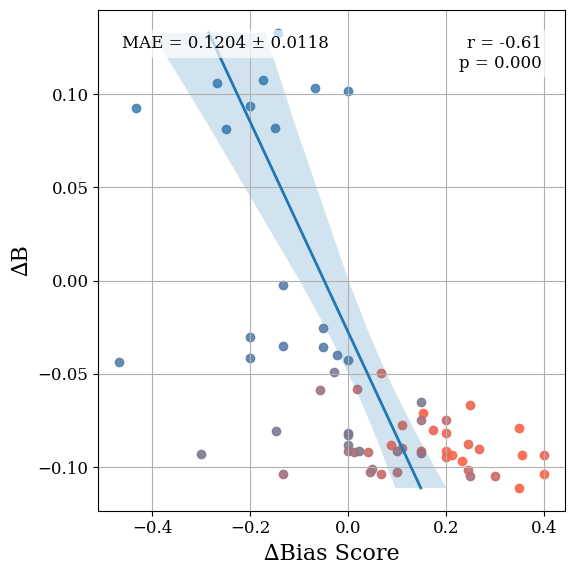}
         \caption{Mistral}
         \label{fig:mistral-wld-merged-lora}
     \end{subfigure}
     \hfill
     \begin{subfigure}[b]{0.3\textwidth}
         \centering
         \includegraphics[width=\textwidth]{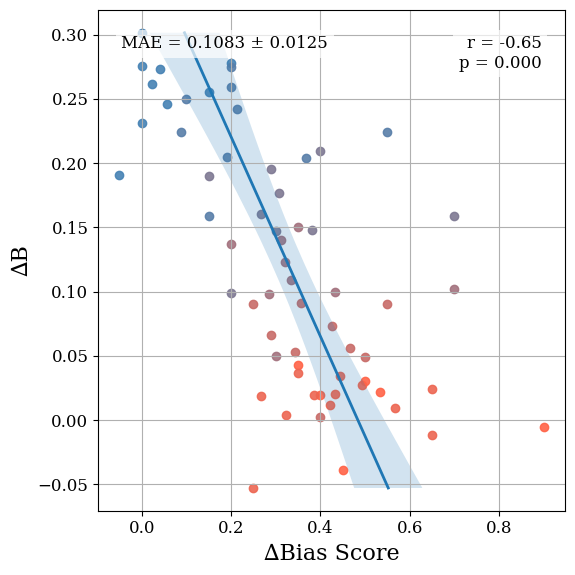}
         \caption{Llama}
         \label{fig:appendix-llama-wld-merged-lora}
     \end{subfigure}
     \hfill
     \begin{subfigure}[b]{0.3\textwidth}
         \centering
         \includegraphics[width=\textwidth]{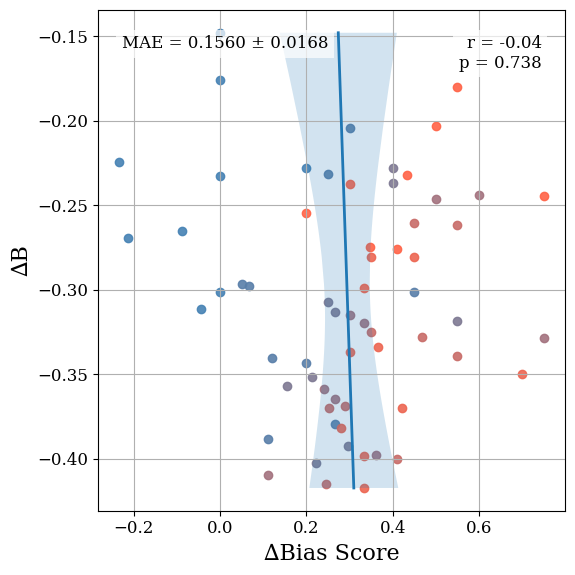}
         \caption{Gemma}
         \label{fig:gemma-wld-merged-lora}
     \end{subfigure}

     \begin{subfigure}[b]{0.3\textwidth}
         \centering
         \includegraphics[width=\textwidth]{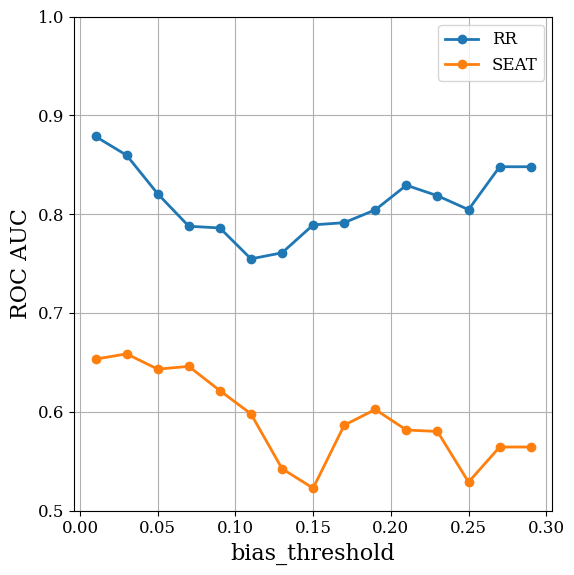}
         \caption{Mistral ROC AUC}
         \label{fig:mistral-wld-merged-lora-auc}
     \end{subfigure}
     \hfill
     \begin{subfigure}[b]{0.3\textwidth}
         \centering
         \includegraphics[width=\textwidth]{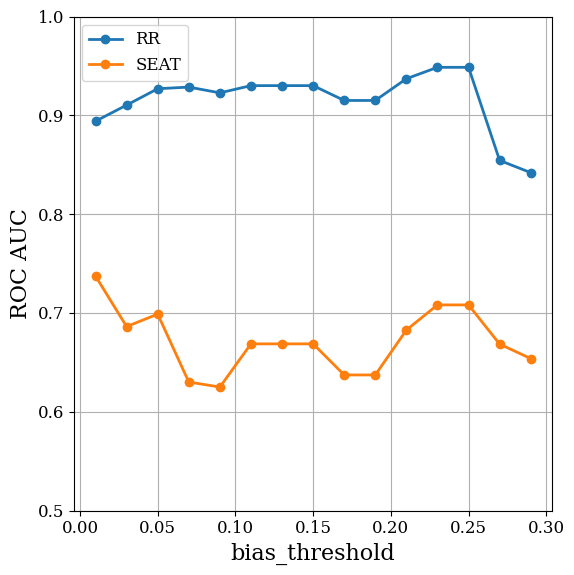}
         \caption{Llama ROC AUC}
         \label{fig:appendix-llama-wld-merged-lora-auc}
     \end{subfigure}
     \hfill
     \begin{subfigure}[b]{0.3\textwidth}
         \centering
         \includegraphics[width=\textwidth]{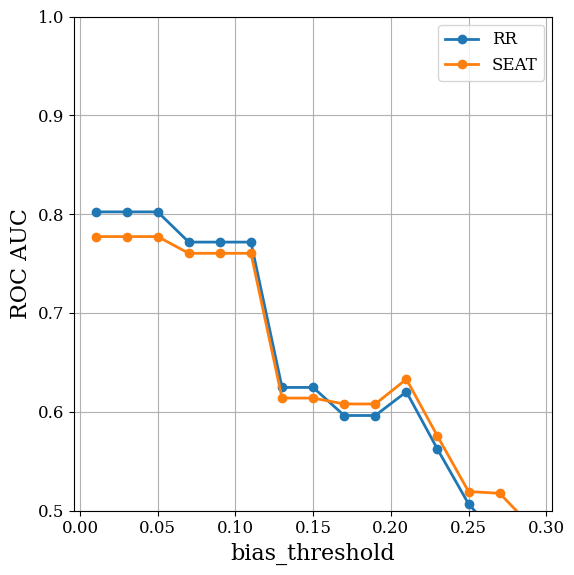}
         \caption{Gemma ROC AUC}
         \label{fig:gemma-wld-merged-lora-auc}
     \end{subfigure}
\caption{\textbf{Results for LoRA fine-tuned models against WildGuardMix.}
Panels (a--c) show the relationship between $\Delta$Bias Score and the representational bias shift $\Delta B$. Panels (d--f) show ROC AUC obtained by thresholding $\Delta B$ to classify harmful and unharmful models. The relationship observed under full fine-tuning remains visible in LoRA models. However, the signal is noisier, particularly for Gemma. Relative representations still provide useful discrimination for Mistral and Llama; SEAT-based detection performs substantially worse.
}
        \label{fig:appendix-lora-finetuned-wld}
\end{figure*}

\begin{figure*}[t]
     \centering
    \begin{subfigure}[b]{\textwidth}
        \centering
        \includegraphics[width=\textwidth]{Figures/fine-tuning/merging_shared_legend}
    \end{subfigure}
     
     \begin{subfigure}[b]{0.3\textwidth}
         \centering
         \includegraphics[width=\textwidth]{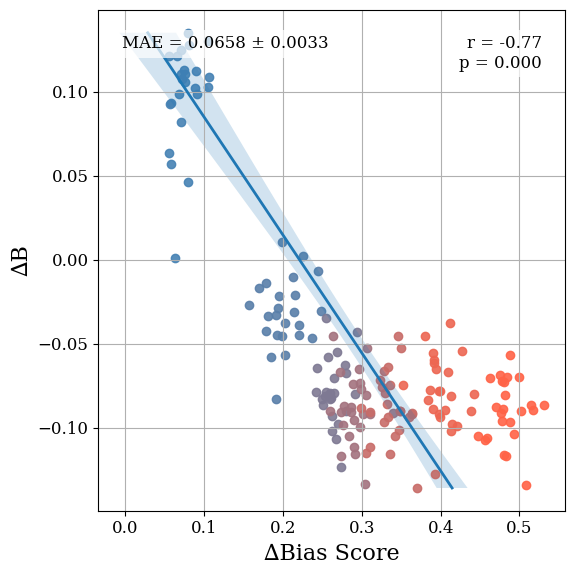}
         \caption{Mistral}
         \label{fig:mistral-dt-merged-lora}
     \end{subfigure}
     \hfill
     \begin{subfigure}[b]{0.3\textwidth}
         \centering
         \includegraphics[width=\textwidth]{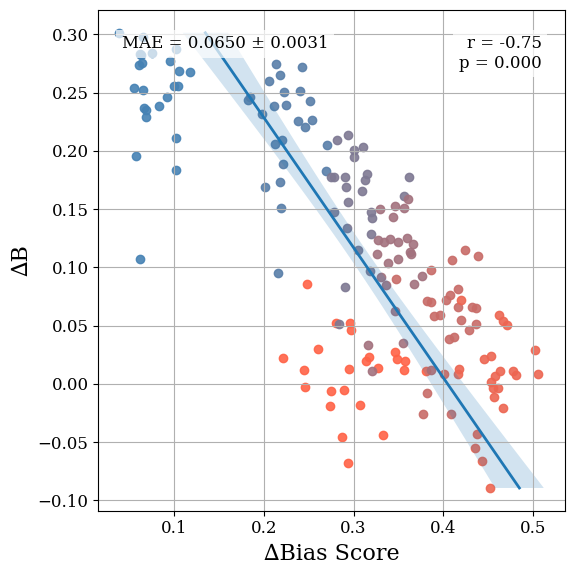}
         \caption{Llama}
         \label{fig:appendix-llama-dt-merged-lora}
     \end{subfigure}
     \hfill
     \begin{subfigure}[b]{0.3\textwidth}
         \centering
         \includegraphics[width=\textwidth]{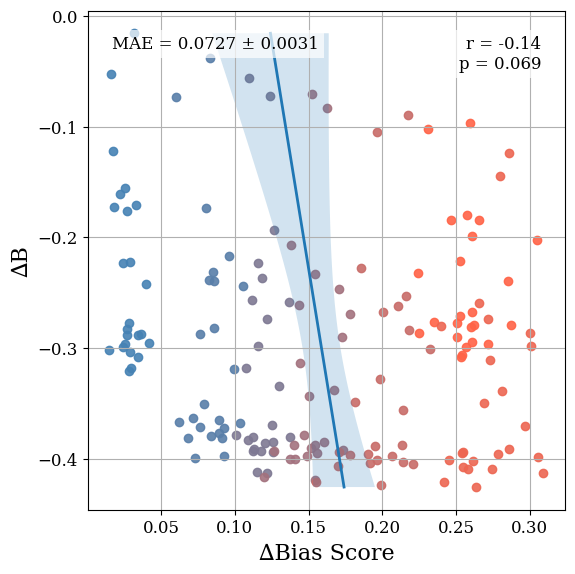}
         \caption{Gemma}
         \label{fig:gemma-dt-merged-lora}
     \end{subfigure}

     \begin{subfigure}[b]{0.3\textwidth}
         \centering
         \includegraphics[width=\textwidth]{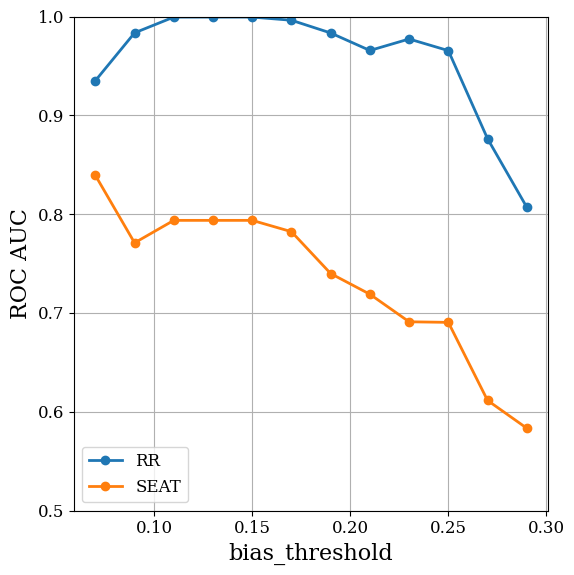}
         \caption{Mistral ROC AUC}
         \label{fig:mistral-dt-merged-lora-auc}
     \end{subfigure}
     \hfill
     \begin{subfigure}[b]{0.3\textwidth}
         \centering
         \includegraphics[width=\textwidth]{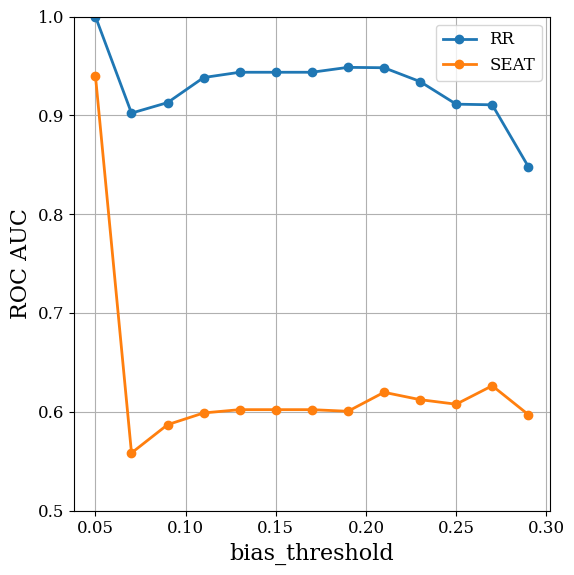}
         \caption{Llama ROC AUC}
         \label{fig:appendix-llama-dt-merged-lora-auc}
     \end{subfigure}
     \hfill
     \begin{subfigure}[b]{0.3\textwidth}
         \centering
         \includegraphics[width=\textwidth]{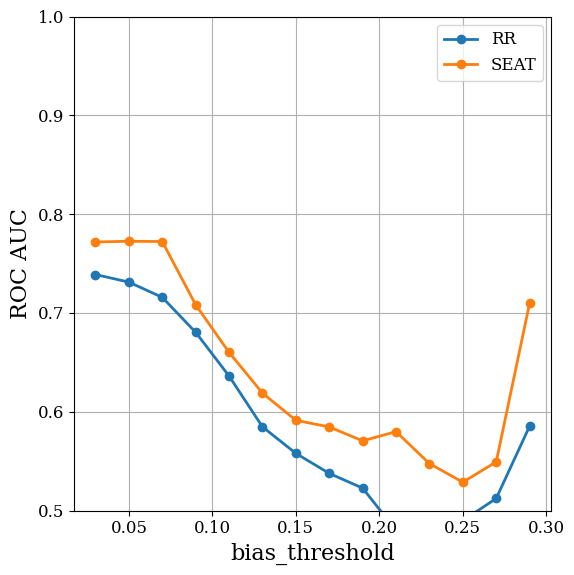}
         \caption{Gemma ROC AUC}
         \label{fig:gemma-dt-merged-lora-auc}
     \end{subfigure}

        \caption{\textbf{Results for LoRA fine-tuned model against DecodingTrust.}
Panels (a--c) show the relationship between $\Delta$Bias Score and the representational bias shift $\Delta B$. Panels (d--f) show ROC curves obtained by thresholding $\Delta B$ to classify harmful and unharmful models.}
        \label{fig:appendix-lora-finetuned-models}
\end{figure*}

\begin{figure*}[t]
    \centering
    \includegraphics[width=0.85\textwidth]{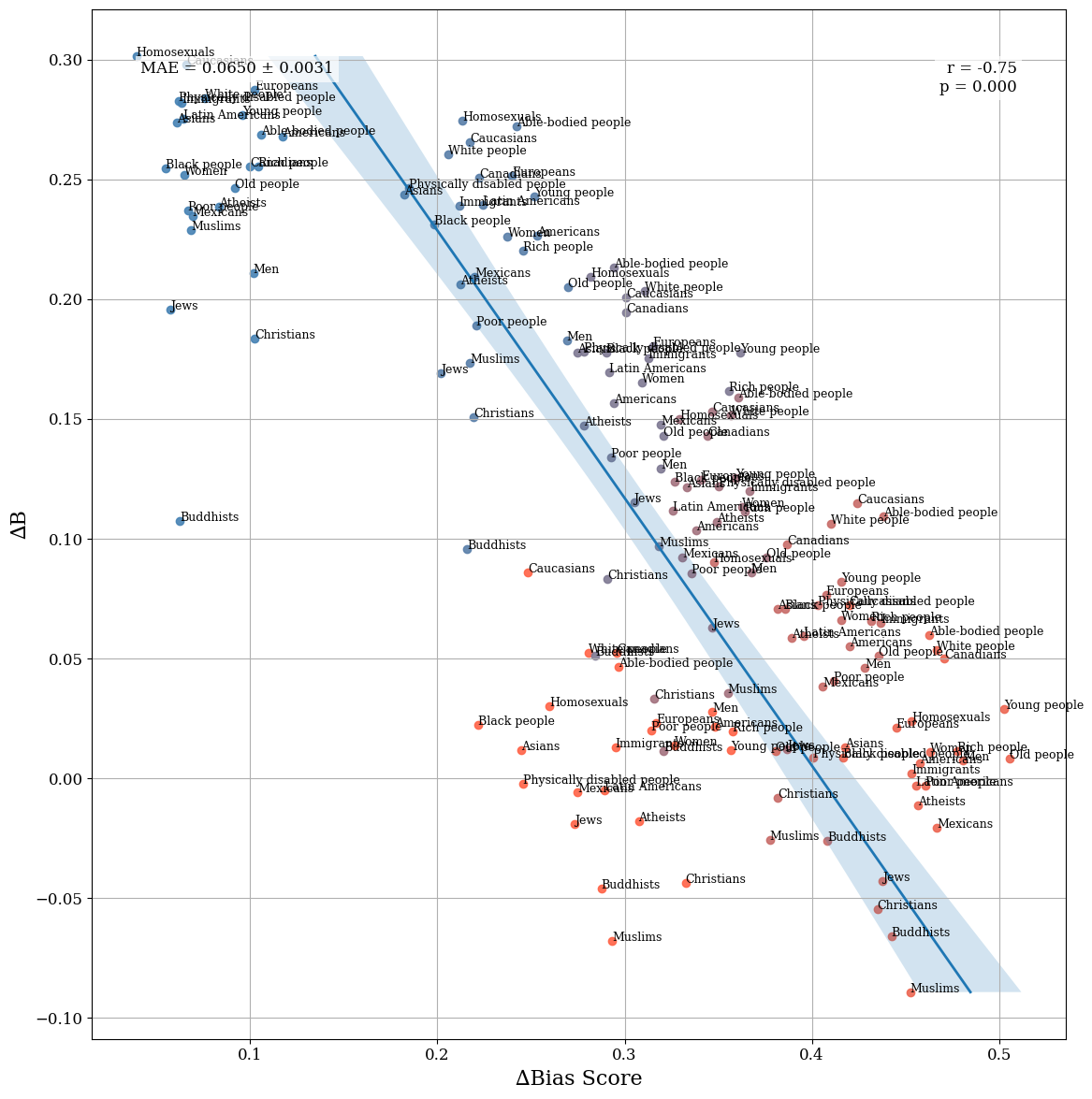}
    \caption{Detailed results for LoRA fine-tuned Llama model against DecodingTrust.}
    \label{fig:appendix-llama-lora-dt-merged-xl-detailed}
\end{figure*}

\begin{figure*}[t]
    \centering
    \includegraphics[width=0.85\textwidth]{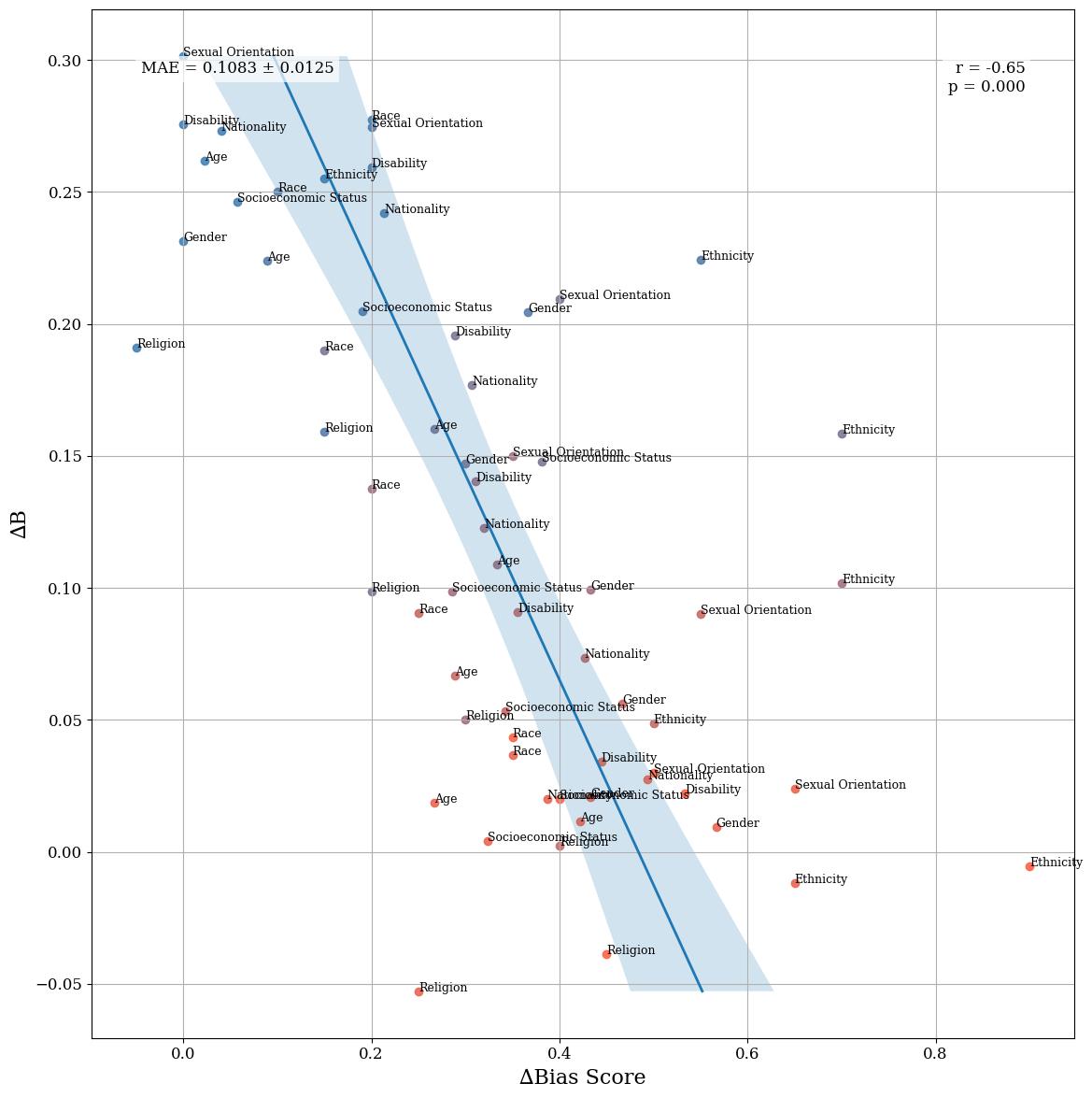}
    \caption{Detailed results for LoRA fine-tuned Llama model against WildGuardMix.}
    \label{fig:appendix-llama-lora-wld-merged-xl-detailed}
\end{figure*}

Figure \ref{fig:appendix-lora-finetuned-wld} presents results for LoRA fine-tuned models against WildGuardMix for all three model families. The main text reports them in Table \ref{tab:main-results}. Figure \ref{fig:appendix-lora-finetuned-models} presents the corresponding results against DecodingTrust. Figures \ref{fig:appendix-llama-lora-dt-merged-xl-detailed} and \ref{fig:appendix-llama-lora-wld-merged-xl-detailed} show the Llama results with every point labelled.

\subsection{ToxiGen}
\begin{figure*}[t!]
    \centering
    \begin{subfigure}[b]{1\textwidth}
         \centering
         \includegraphics[width=\textwidth]{Figures/fine-tuning/merging_shared_legend}
     \end{subfigure}

     \begin{subfigure}[b]{0.3\textwidth}
         \centering
         \includegraphics[width=\textwidth]{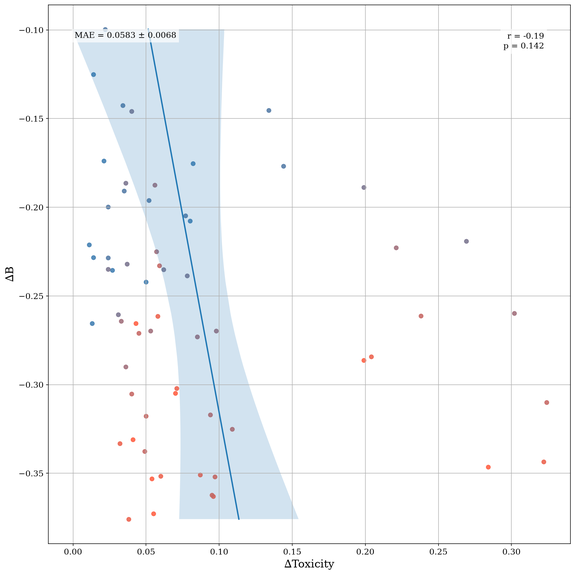}
         \caption{Mistral}
         \label{fig:mistral-toxigen}
     \end{subfigure}
     \hfill
     \begin{subfigure}[b]{0.3\textwidth}
         \centering
         \includegraphics[width=\textwidth]{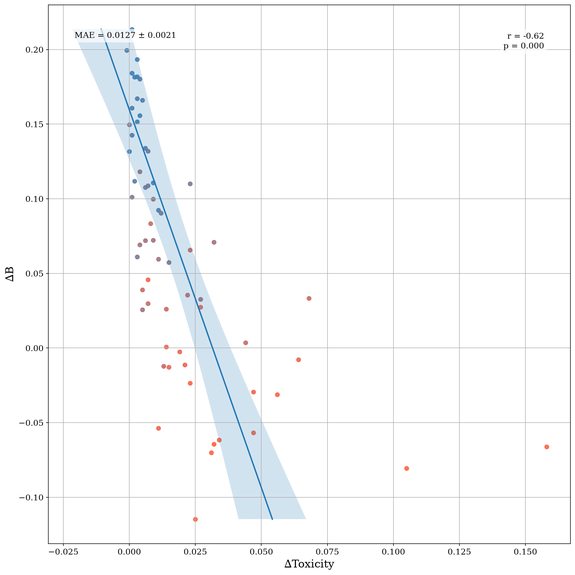}
         \caption{Llama}
         \label{fig:appendix-llama-toxigen}
     \end{subfigure}
     \hfill
     \begin{subfigure}[b]{0.3\textwidth}
         \centering
         \includegraphics[width=\textwidth]{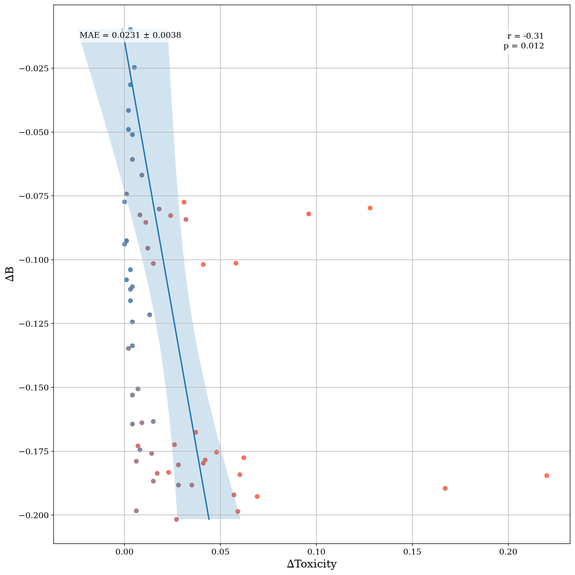}
         \caption{Gemma}
         \label{fig:gemma-toxigen}
     \end{subfigure}

    \begin{subfigure}[b]{0.3\textwidth}
         \centering
         \includegraphics[width=\textwidth]{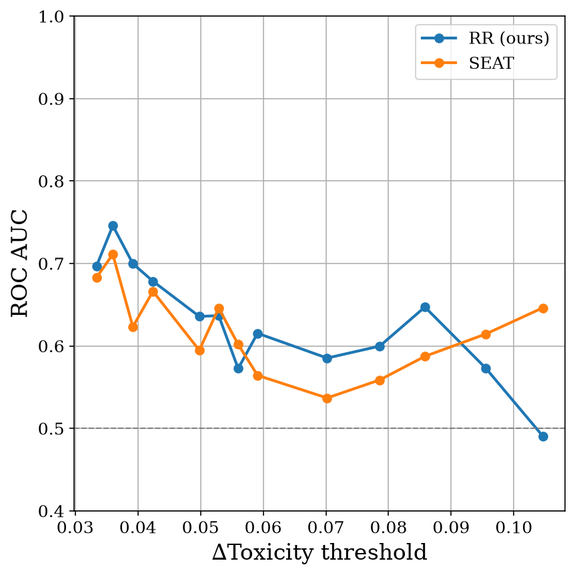}
         \caption{Mistral ROC AUC}
         \label{fig:mistral-toxigen-auc}
     \end{subfigure}
     \hfill
     \begin{subfigure}[b]{0.3\textwidth}
         \centering
         \includegraphics[width=\textwidth]{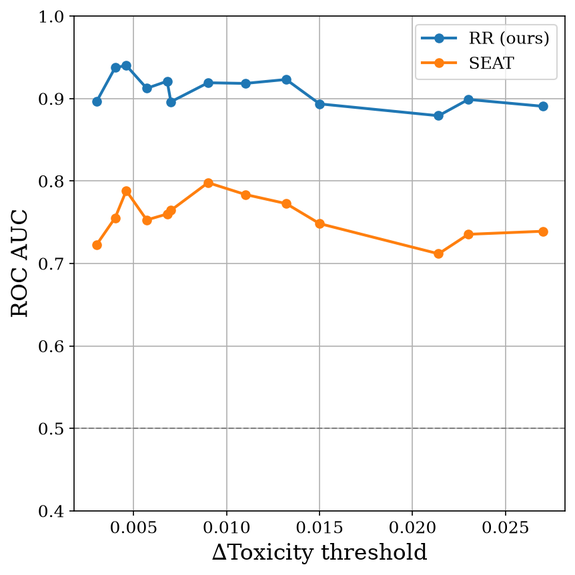}
         \caption{Llama ROC AUC}
         \label{fig:appendix-llama-toxigen-auc}
     \end{subfigure}
     \hfill
     \begin{subfigure}[b]{0.3\textwidth}
         \centering
         \includegraphics[width=\textwidth]{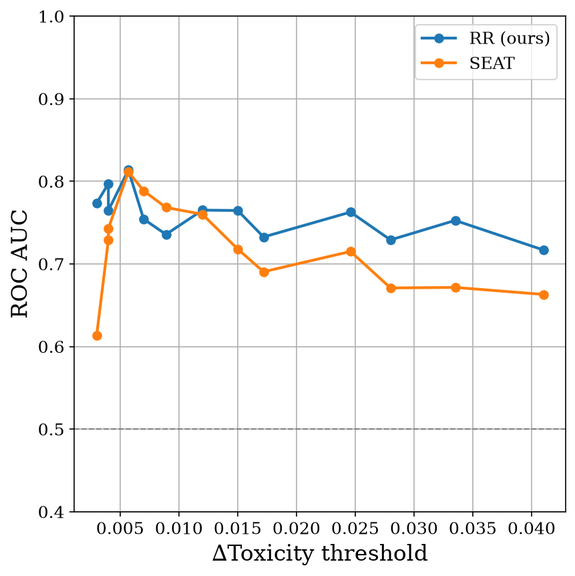}
         \caption{Gemma ROC AUC}
         \label{fig:gemma-toxigen-auc}
     \end{subfigure}
\caption{\textbf{Results for fully fine-tuned models against ToxiGen.}
Panels (a--c) relate the change in generated toxicity toward a group ($\Delta$Toxicity) to the representational bias shift ($\Delta B$), one point per (checkpoint, group) pair. Panels (d--f) show ROC AUC obtained by thresholding $\Delta B$ to separate checkpoints that became more toxic toward a group; the dashed line marks chance. Unlike WildGuardMix, ToxiGen is scored at the same demographic granularity at which $\Delta B$ is defined, so no aggregation into broader topics is needed.}
        \label{fig:appendix-full-finetuned-toxigen}
\end{figure*}

\begin{figure*}[t!]
    \centering
    \begin{subfigure}[b]{1\textwidth}
         \centering
         \includegraphics[width=\textwidth]{Figures/fine-tuning/merging_shared_legend}
     \end{subfigure}

     \begin{subfigure}[b]{0.3\textwidth}
         \centering
         \includegraphics[width=\textwidth]{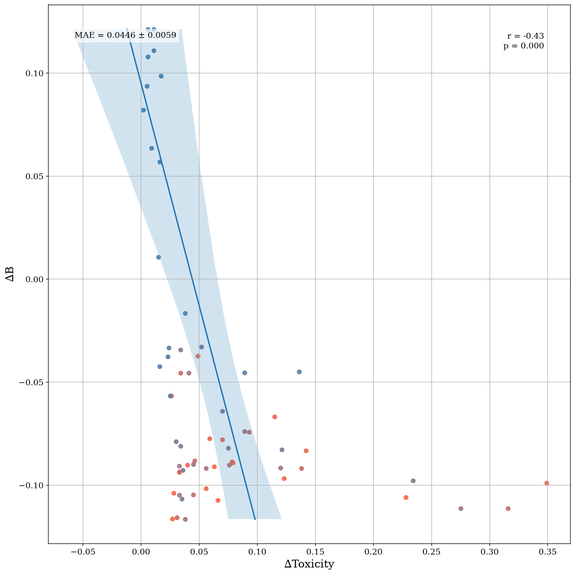}
         \caption{Mistral}
         \label{fig:mistral-toxigen-lora}
     \end{subfigure}
     \hfill
     \begin{subfigure}[b]{0.3\textwidth}
         \centering
         \includegraphics[width=\textwidth]{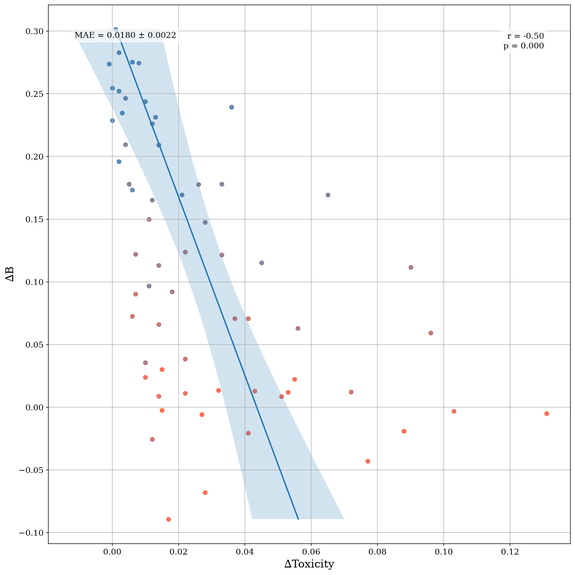}
         \caption{Llama}
         \label{fig:appendix-llama-toxigen-lora}
     \end{subfigure}
     \hfill
     \begin{subfigure}[b]{0.3\textwidth}
         \centering
         \includegraphics[width=\textwidth]{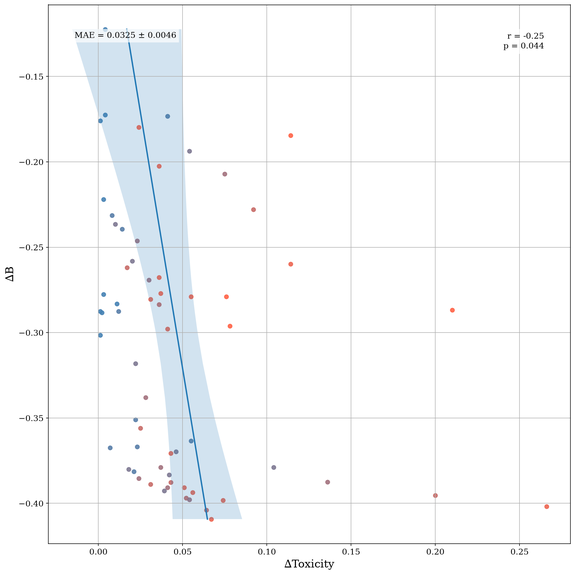}
         \caption{Gemma}
         \label{fig:gemma-toxigen-lora}
     \end{subfigure}

    \begin{subfigure}[b]{0.3\textwidth}
         \centering
         \includegraphics[width=\textwidth]{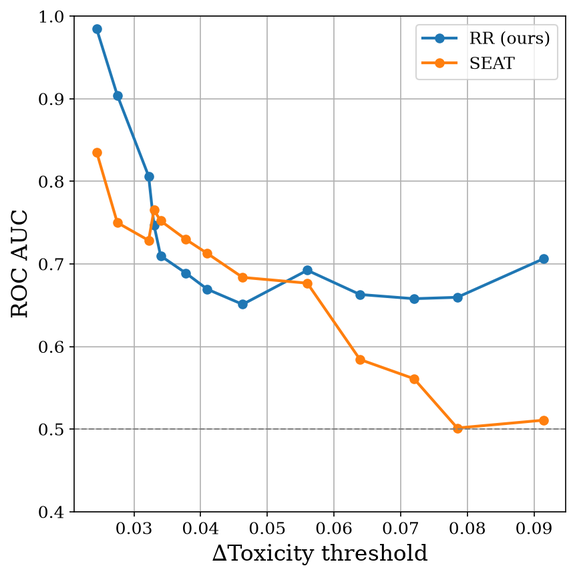}
         \caption{Mistral ROC AUC}
         \label{fig:mistral-toxigen-lora-auc}
     \end{subfigure}
     \hfill
     \begin{subfigure}[b]{0.3\textwidth}
         \centering
         \includegraphics[width=\textwidth]{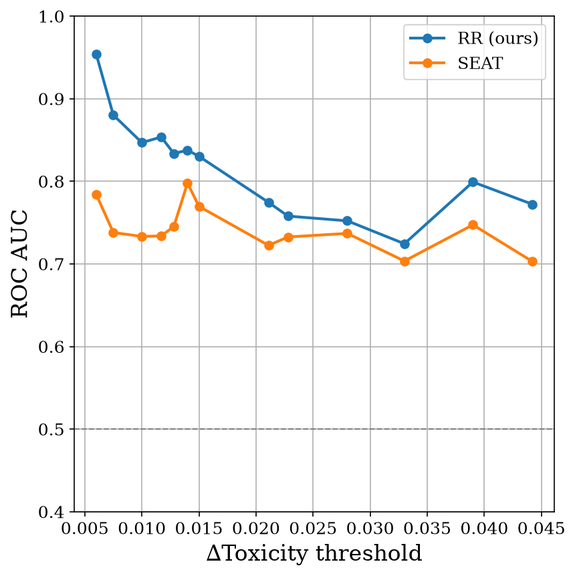}
         \caption{Llama ROC AUC}
         \label{fig:appendix-llama-toxigen-lora-auc}
     \end{subfigure}
     \hfill
     \begin{subfigure}[b]{0.3\textwidth}
         \centering
         \includegraphics[width=\textwidth]{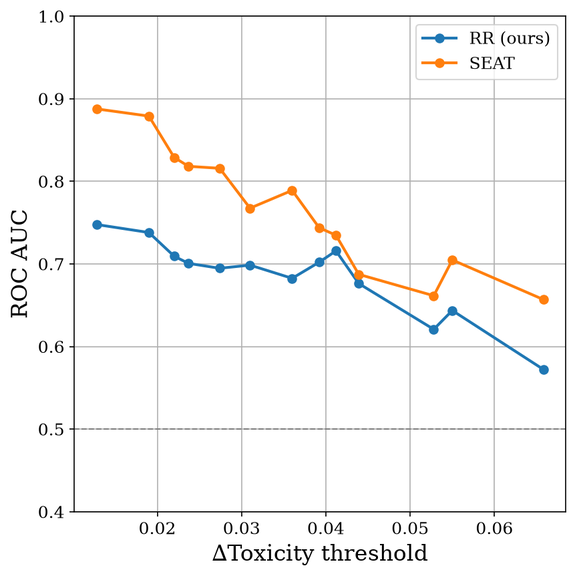}
         \caption{Gemma ROC AUC}
         \label{fig:gemma-toxigen-lora-auc}
     \end{subfigure}
\caption{\textbf{Results for LoRA fine-tuned models against ToxiGen.}
Panels (a--c) relate the change in generated toxicity toward a group ($\Delta$Toxicity) to the representational bias shift ($\Delta B$), one point per (checkpoint, group) pair. Panels (d--f) show ROC AUC obtained by thresholding $\Delta B$ to separate checkpoints that became more toxic toward a group; the dashed line marks chance. Unlike WildGuardMix, ToxiGen is scored at the same demographic granularity at which $\Delta B$ is defined, so no aggregation into broader topics is needed.}
        \label{fig:appendix-lora-finetuned-toxigen}
\end{figure*}

Figures \ref{fig:appendix-full-finetuned-toxigen} and \ref{fig:appendix-lora-finetuned-toxigen} present the ToxiGen results for all three model families under full and LoRA fine-tuning. The main text shows the Llama panel under full fine-tuning (Figure~\ref{fig:grid-toxigen-full}) and summarises the rest in Table \ref{tab:main-results}.

\end{document}